\PassOptionsToPackage{dvipsnames,svgnames,x11names}{xcolor}
\documentclass[10pt,twocolumn,letterpaper]{article}

\usepackage{config/iccv}
\usepackage{config/iccv23}
\newcommand{\NOBJ}{81\xspace}
\newcommand{\NSUBJ}{46\xspace}
\newcommand{\NINST}{32\xspace}
\newcommand{\NSEQ}{1390\xspace}

\newcommand{\NHOURS}{17.3\xspace}

\newcommand{\LL}{\mathcal{L}}

\newcommand{\OO}{\mathcal{O}}

\newcommand{\specialcell}[2][c]{%
\begin{tabular}[#1]{@{}c@{}}#2\end{tabular}}
\newcolumntype{x}{>{\columncolor{MistyRose}}c}
\newcolumntype{y}{>{\columncolor{LightCyan1}}c}

\acrodef{hoi}[HOI]{Human-Object Interaction}
\acrodef{ahoi}[AHOI]{Articulated Human-Object Interaction}
\acrodef{fahoi}[f-AHOI]{Full-Body Articulated Human-Object Interaction}
\acrodef{dataset}[\texttt{CHAIRS}]{Capturing Human and Articulated-object InteRactionS}
\acrodef{cvae}[cVAE]{conditional Variational Auto-Encoder}
\acrodef{icp}[ICP]{Iterative Closest Points}
\acrodef{tlcc}[TLCC]{time-lagged cross-correlation}
\acrodef{urdf}[URDF]{Unified Robot Description Format}

\definecolor{optyellow}{HTML}{BFAF6E}
\definecolor{optblue}{HTML}{7480E7}
\iccvfinalcopy
\ificcvfinal\fi

\begin{document}
\title{Full-Body Articulated Human-Object Interaction}
\author{
Nan Jiang$^{1,2\,\star{}\,\dagger}$, Tengyu Liu$^{2\,\star{}}$, Zhexuan Cao$^{3\,\dagger}$, Jieming Cui$^{1,2,\dagger}$, Zhiyuan Zhang$^{2,3,\dagger}$,\\Yixin Chen$^2$, He Wang$^4$, Yixin Zhu$^{5\,\textrm{\Letter}}$, Siyuan Huang$^{2\,\textrm{\Letter}}$
\vspace{0.5em}\\
\hspace{-12pt}\begin{tabular}{r l}
\small\url{https://jnnan.github.io/project/chairs/}&
\small$\textrm{\Letter}$\,\,\texttt{yixin.zhu@pku.edu.cn,\,syhuang@bigai.ai}\\
\small$^\star{}$Equal contributors\quad{}
\small$^\dagger$ Work done during an internship at BIGAI&
\small$^1$ School of Intelligence Science and Technology, Peking University\\
\small$^2$ Beijing Institute of General Artificial Intelligence (BIGAI)&
\small$^3$ Department of Automation, Tsinghua University\\
\small$^4$ Center on Frontiers of Computing Studies, Peking University&
\small$^5$ Institute for Artificial Intelligence, Peking University
\end{tabular}
\vspace{-15pt}
}

\twocolumn[{%
\renewcommand\twocolumn[1][]{#1}%
    \maketitle
    \centering
    \captionsetup{type=figure}
    \includegraphics[width=\linewidth]{teaser}
    \captionof{figure}{\textbf{Examples of the proposed \acf{dataset} dataset.} It contains fine-grained interactions between \NSUBJ participants and \NOBJ sittable objects with drastically different kinematic structures, providing multi-view RGB-D sequences and ground-truth 3D mesh of humans and articulated objects for over \NHOURS hours of recordings.\vspace{12pt}}
    \label{fig:teaser}
}]

\begin{abstract}\vspace{-6pt}
Fine-grained capture of 3D \acp{hoi} enhances human activity comprehension and supports various downstream visual tasks. However, previous models often assume that humans interact with rigid objects using only a few body parts, constraining their applicability. In this paper, we address the intricate challenge of \ac{fahoi}, where complete human bodies interact with articulated objects having interconnected movable joints. We introduce \ac{dataset}, an extensive motion-captured \ac{fahoi} dataset comprising \NHOURS hours of diverse interactions involving \NSUBJ participants and \NOBJ articulated as well as rigid sittable objects. The \ac{dataset} provides 3D meshes of both humans and articulated objects throughout the interactive sequences, offering \textbf{realistic} and \textbf{physically plausible} full-body interactions. We demonstrate the utility of \ac{dataset} through object pose estimation. Leveraging the geometric relationships inherent in \ac{hoi}, we propose a pioneering model that employs human pose estimation to address articulated object pose and shape estimation within whole-body interactions. Given an image and an estimated human pose, our model reconstructs the object's pose and shape, refining the reconstruction based on a learned interaction prior. Across two evaluation scenarios, our model significantly outperforms baseline methods. Additionally, we showcase the significance of \ac{dataset} in a downstream task involving human pose generation conditioned on interacting with articulated objects. We anticipate that the availability of \ac{dataset} will advance the community's understanding of finer-grained interactions.
\end{abstract}

\begin{table*}[t!]
    \centering
    \small
    \setlength{\tabcolsep}{3pt}
    \caption{\textbf{Comparisons between \acs{dataset} and other \acs{hoi} datasets.}}
    \resizebox{\linewidth}{!}{%
        \begin{tabular}{cccccccccc}
            \toprule
            Dataset & \# object & \# participants & \# instructions & \# hours & fps & \# view & articulated objects & human & annotation type \\
            \midrule
            GRAB~\cite{taheri2020grab}  & 51 & 10 & 4 & 3.8 & 120 & 0 & No & Whole-body & mocap\\
            D3D-HOI~\cite{xu2021d3d}  & 24 & 5 & / & 0.6 & 3 & 1 & Yes & Whole-body & manual \\
            BEHAVE~\cite{bhatnagar2022behave} & 20 & 8 & 6 & 4.2 & 30 & 4 & No & Whole-body & multi-kinect\\
            ARCTIC~\cite{fan2022articulated} & 10 & 9 & 1 & 1.2 & 30 & 8+1 & Yes & Two hands & mocap\\
            COUCH~\cite{zhang2022couch} & 4 & 6 & 6 & 3 & 60 & 4 & No & Whole-body & mocap \\
            \ac{dataset} (Ours) & \NOBJ & \NSUBJ & \NINST & \NHOURS & 30 & 4 & Yes & Whole-body & mocap\\
            \bottomrule
        \end{tabular}%
    }%
    \label{tab:dataset}
\end{table*}

\section{Introduction}

In the realm of computer vision and robotics, the fundamental comprehension of \acf{hoi}~\cite{liao2020ppdm,liao2022gen,zhang2021mining,yuan2022detecting} lies at the core of dissecting intricate human activities. This paper embarks on unraveling the complex challenge of \acf{fahoi}. This endeavor mandates tackling two pivotal dimensions: (i) fashioning kinematic-agnostic representations for \textbf{articulated} objects and (ii) delving into the intricate spatial-temporal tapestry interweaving objects with human \textbf{whole-bodies}. Our primary focus resides in addressing the intricate task of object pose estimation within the realm of \ac{fahoi}, considering that reconstructing 3D human poses from frontal viewpoints is comparably uncomplicated.

The crux of object pose estimation within the context of \ac{fahoi} is punctuated by three principal challenges:

\paragraph{The dearth of comprehensive \ac{fahoi} datasets}

Existing strides in 3D \ac{hoi} predominantly either assume interactions with rigid objects or confine themselves to involving specific segments of the human anatomy~\cite{bhatnagar2022behave,taheri2020grab,zhang2020perceiving,li2020detailed,zhang2022couch,fan2022articulated,xu2021d3d,haresh2022articulated}. Regrettably, these assumptions drastically oversimplify genuine human interactions that span diverse body parts engaging with articulated objects embodying moveable elements such as cabinets and office chairs. A more intricate level of interaction necessitates a richer dataset.

\paragraph{The multifaceted landscape of object kinematic structures}

Objects constituting the realm of \ac{fahoi} exhibit notable disparities in their kinematic frameworks, even when categorized under the same umbrella. Prevailing methodologies often lean towards uniform structures~\cite{xu2021d3d,liu2022akb,fan2022articulated,haresh2022articulated}, thereby disregarding the diverse tapestry that constitutes real-world scenarios. The endeavor of accurately reconstructing objects manifesting divergent geometries and structures is plagued with its own set of challenges.

\paragraph{The intricacies of complex interactions}

Engaging with articulated objects entails grappling with intricate spatial and physical relationships, often entailing occlusions and intricate points of contact. The intricacy of these dynamics thrusts conventional pose estimation mechanisms reliant on point cloud template-matching~\cite{zhang2020perceiving,wang2019normalized,he2020pvn3d,peng2019pvnet,li2020category} into the realm of insufficiency. The prominence of contacts further compounds the endeavor of precise reconstruction, as slight inaccuracies can swiftly usher implausible interactions into the picture.

The trajectory of this research endeavors to navigate the above three challenges through the prism of three principal solutions, respectively:

To confront the scarcity of \ac{fahoi} datasets, we introduce \ac{dataset}, a multi-view RGB-D dataset. Illustrated in \cref{fig:teaser}, \ac{dataset} chronicles a diverse tapestry of interactions, seamlessly intertwining \NSUBJ participants with \NOBJ sittable objects (\eg, chairs, sofas, stools, and benches). 28 of these objects are endowed with moveable parts. Each frame encapsulates 3D meshes of both human \textbf{whole-bodies} and objects, casting a spotlight on interactions with sittable objects that encompass a diverse spectrum of structures and distinctive movable elements conducive to multifarious human interactions.

To traverse the labyrinth of kinematic diversity, \ac{dataset} meticulously selects representative objects characterized by an eclectic array of structures. Unlike traditional datasets and methodologies tethered to uniform kinematics~\cite{xiang2020sapien,wang2019shape2motion,liu2022akb}, we champion real-world heterogeneity, encompassing the gamut from rigid stools to swivel chairs boasting up to 7 movable components. Each component is linked to its parent through a nexus of revolute, prismatic, or composite joints.

To unravel the enigma of complex interactions, we proffer an innovative approach to articulated object pose estimation, one that harnesses the subtle interplay of fine-grained interaction relationships to reconstruct the object in question. This approach diverges from the conventional recourse of manually labeling contact maps corresponding to human body parts~\cite{zhang2020perceiving,hassan2019resolving,bhatnagar2022behave}. Instead, our approach melds the intricacies of these relationships with a reconstruction model and an interaction prior, the latter of which is imbued with the essence of a \ac{cvae}. This evolution sidesteps the need for predefined knowledge grounded in laborious annotation. Moreover, the significance of these intricate relationships is showcased through our venture into learning human poses within the ambit of articulated objects. By juxtaposing the generative prowess harnessed from \ac{dataset} with that stemming from a dataset centered on rigid objects~\cite{zhang2022couch}, we underscore the pivotal role played by the nuanced geometrical relationships encapsulated within \ac{dataset} in the broader canvas of downstream tasks.

Our \textbf{contributions} are four-fold:
(i) \ac{dataset}, a sprawling multi-view RGB-D repository infused with diverse 3D meshes.
(ii) The seamless extension of articulated object pose estimation to the arduous landscape of \ac{fahoi}.
(iii) An object pose estimation approach that transcends the strictures of structure.
(iv) An overarching interaction prior that captures the subtleties of fine-grained interactions, acting as a catalyst for the journey of pose estimation.

\begin{figure*}[t!]
    \centering
    \begin{subfigure}{0.5\linewidth}
        \includegraphics[width=\linewidth]{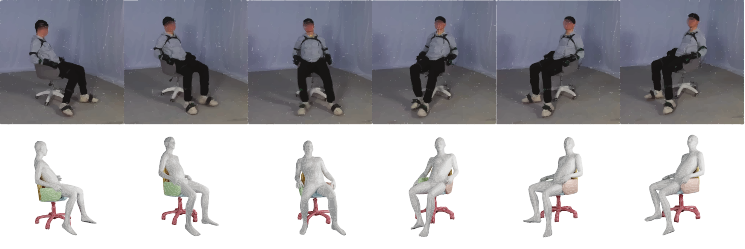}%
        \\
        \includegraphics[width=\linewidth]{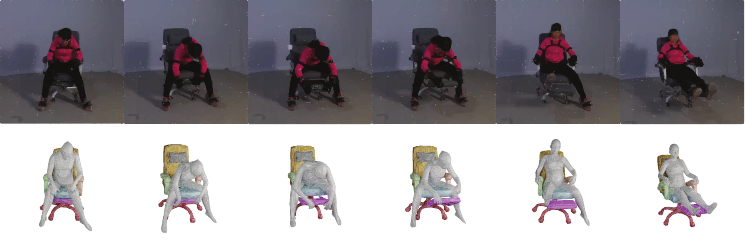}%
        \caption{sequences of objects articulating over time}
    \end{subfigure}%
    \hfill
    \begin{subfigure}{0.5\linewidth}
        \includegraphics[width=\linewidth]{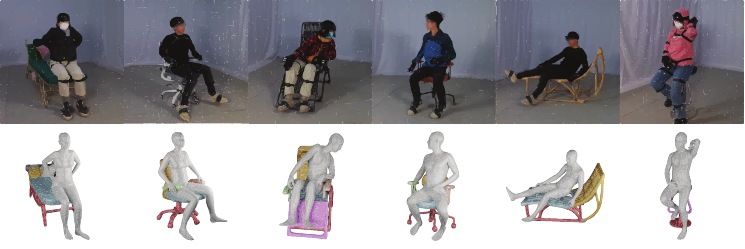}%
        \\
        \includegraphics[width=\linewidth]{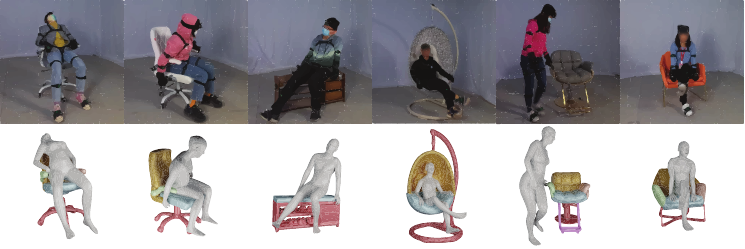}%
        \caption{frames of diverse interactions}
    \end{subfigure}%
    \caption{\textbf{Samples from the \ac{dataset} dataset.} \ac{dataset} encompasses diverse \acsp{ahoi} captured through precisely calibrated multi-view RGB-D cameras, offering detailed 3D meshes of human participants and articulated objects. The figure showcases (a) RGB frames alongside corresponding ground-truth mesh sequences and (b) an array of varied \acs{ahoi} instances.}
    \label{fig:dataset}
\end{figure*}

\section{Related Work}

\paragraph{3D \acf{hoi}}

The evolution of \ac{hoi} research spans from 2D image-based interaction detection~\cite{chao2015hico,qi2018learning,gkioxari2018detecting,liao2020ppdm,liao2022gen,zhang2021mining,yuan2022detecting,huang2023diffusion} to 3D interaction reconstruction~\cite{savva2016pigraphs,hassan2019resolving,chen2019holistic,weng2021holistic,xu2021d3d,zanfir2018monocular,siwei2021learning} and generation~\cite{hassan2021populating,wang2021synthesizing,xu2020hierarchical,holden2017phase,wang2022humanise,zhang2022couch} within 3D scenes. Notably, PiGraph~\cite{savva2016pigraphs} and D3D-HOI~\cite{hassan2019resolving,xu2021d3d} capture daily activities and reconstruct interactions, often relying on visual observations. In contrast, MoCap systems~\cite{taheri2020grab,bhatnagar2022behave,fan2022articulated,zhang2022couch} offer fine-grained 3D human-object interactions. GRAB~\cite{taheri2020grab} and ARCTIC~\cite{fan2022articulated} emphasize interactions with small objects, while BEHAVE~\cite{bhatnagar2022behave} and COUCH~\cite{zhang2022couch} involve interactions with everyday objects. However, these works often focus on rigid objects or hand-object interactions with articulated objects. In contrast, our \ac{dataset} dataset captures realistic \textit{whole-body} interactions with diverse articulated objects.

\paragraph{Articulated Human-Object Interaction}

\acp{ahoi} build on part-level object representations, modeling intricate spatial-temporal interactions between humans and articulated objects~\cite{haresh2022articulated}. Noteworthy contributions include D3D-HOI~\cite{xu2021d3d}, ARCTIC~\cite{fan2022articulated}, and 3DADN~\cite{qian2022understanding}. D3D-HOI~\cite{xu2021d3d} captures humans interacting with containers, ARCTIC~\cite{fan2022articulated} focuses on motion-captured RGB-D hand-object interactions, and 3DADN~\cite{qian2022understanding} annotates movable object parts from internet videos. However, these works often emphasize hand-object interactions, whereas our focus extends to \acp{ahoi} encompassing diverse articulated objects and multiple body parts.

\paragraph{Contact-Rich \ac{hoi}}

The realm of \ac{fahoi} requires a deeper \ac{hoi} understanding. While 3D \ac{hoi} literature has expanded, few works address full-body contacts through reconstruction~\cite{hassan2019resolving} or generation~\cite{zhao2022compositional,wang2022humanise,hassan2021stochastic,zhang2022couch}. However, these works often focus on static scenes with limited interactions. In contrast, our \ac{dataset} dataset encompasses diverse articulated objects and interactions.

\paragraph{Articulated Object Pose Estimation}

The estimation of 6-DOF poses for rigid objects has garnered attention~\cite{kang2020yolo,he2020pvn3d,braun2016pose,wang2019densefuison,peng2019pvnet,park2019pix2pose,do2018deep}. Template-based methods~\cite{hinterstoisser2011multimodal,yang2015go,wang2019normalized,kehl2017ssd} and regression models~\cite{abbatematteo2019learning} are common, with recent strides in articulated object pose estimation~\cite{desingh2019factored,michel2015pose,li2020category} leveraging these techniques. Regression and implicit function models~\cite{mu2021a,tseng2022cla,yang2021lasr,jiang2022ditto} are also explored. Despite progress, these methods often assume consistent kinematic structures within object categories. In contrast, our \ac{dataset} dataset features diverse kinematic structures and models capable of handling various parts and kinematics of 3D objects.

\section{The \texorpdfstring{\texttt{CHAIRS}}{} Dataset}

A significant challenge in modeling \acp{ahoi} is the lack of accurate 3D annotations. To address this gap, we introduce \ac{dataset}, a comprehensive \ac{ahoi} dataset featuring multi-view RGB-D sequences. \ac{dataset} offers precise 3D meshes of humans and articulated objects during interactions, captured through a hybrid motion capture (MoCap) system that combines inertial and optical tracking techniques. The data collection process prioritizes realism and physical authenticity, resulting in a dataset that significantly advances interaction understanding. A detailed comparison between \ac{dataset} and previous \ac{hoi} datasets is outlined in \cref{tab:dataset}.

\subsection{Data Collection}

\paragraph{Overview}

\ac{dataset} encompasses a total of \NSEQ sequences depicting articulated interactions involving humans and sittable objects like chairs, sofas, stools, and benches. Exemplary sequences from \ac{dataset} and a showcase of the object variety can be seen in \cref{fig:dataset}. Each object's exploration involves 6 distinct participants, each contributing three interaction sequences, resulting in 18 sequences for each object. In every sequence, participants execute 6 diverse actions drawn randomly from a pool of \NINST interactions, such as shifting a stool, reclining on a sofa, or rotating a chair; please refer to the Supplementary Material ~\cref{supp:instructions} for further details. Participant instructions were kept deliberately high-level to ensure authentic and natural performances.

\paragraph{Object diversity}

The object gallery in \ac{dataset} boasts an array of objects, each possessing a range of appearances and kinematic structures. Objects were curated by sourcing them online, with a focus on maximizing stylistic diversity. Notably, 28 objects incorporate at least one articulated joint, contributing to rich interaction scenarios. The 3D meshes of these objects were captured using the Scaniverse app on an iPad Pro (11-inch, 2nd generation) and subsequently refined manually to eliminate any imperfections. The 3D meshes were further segmented using the annotation tool~\cite{mo2019partnet} into eight functional parts. Participants received context-specific instructions tailored to the object they were interacting with.

\paragraph{Camera and hardware setup}

\begin{figure}[t!]
    \centering
    \includegraphics[width=\linewidth]{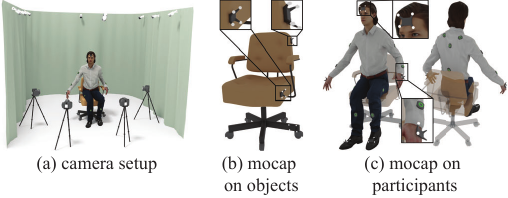}
    \caption{\textbf{Setup for data collection in the \ac{dataset} dataset.} Our data collection setup consisted of (a) four front-facing RGB-D cameras supplemented by a network of motion capture cameras surrounding the recording area, (b) hybrid trackers affixed to various movable parts of objects, and (c) a configuration incorporating five hybrid trackers and seventeen IMUs distributed on the participants.}
    \label{fig:setup}
\end{figure}

As depicted in \cref{fig:setup}, all sequences were exclusively captured in a controlled laboratory setup, encompassing a designated area of 5m$\times$4m ensuring complete visibility of all actions for the cameras. Four Kinect Azure DK cameras, strategically positioned to capture front-facing multi-view perspectives, were employed to record the interactions. These cameras were meticulously calibrated and synchronized. To ensure the precision of ground-truth poses for both humans and objects, a commercial inertial-optical hybrid MoCap system was incorporated alongside the Kinect setup; for further specifics, see the subsequent section.

\subsection{Motion Capture (MoCap) System}

\paragraph{Hybrid MoCap}

Our MoCap system is composed of a MoCap suit outfitted with 5 hybrid trackers and 17 wearable Inertial Measurement Units (IMUs), alongside a pair of gloves equipped with 12 IMUs each. The setup further includes supplementary hybrid trackers and a collection of 8 high-speed cameras. A hybrid tracker, which encompasses 4 optical markers and an IMU, is capable of accurately measuring its own 6D pose even in conditions of substantial occlusion. The arrangement of our data collection setup is illustrated in \cref{fig:setup}. For capturing the pose of a human or an object part, either an IMU or a hybrid tracker can be utilized to record the global orientation or 6D pose, respectively.

\paragraph{Capturing articulated object poses}

The recording process of articulated object poses in the context of interactions unfolds across three phases. First, we position the object into its canonical pose and affix a hybrid tracker to each of its movable components. Subsequently, we calculate the relative transformation between the object part and the trackers. During the recording process, the real-time ground-truth 6D pose of each object part is computed based on the tracker poses. Finally, we match the rigid parts to the kinematic structure of the object to yield high-fidelity object poses.

\paragraph{Capturing human body poses}

For human poses and shapes, we adopt the SMPL-X~\cite{pavlakos2019expressive} representation. Participants are attired in a MoCap suit incorporating 17 IMUs, don a pair of MoCap gloves, and have 5 hybrid trackers affixed to their heads, hands, and feet. It is noteworthy that while hybrid trackers capture 6D poses, IMUs solely measure global orientations. The optimization of human model shape parameters is undertaken to ensure that the reconstructed SMPL-X mesh aligns with the positions of the hybrid trackers. As a result, the MoCap system delivers real-time estimated human poses and shapes during recording.

\begin{figure}[b!]
    \centering
    \includegraphics[width=\linewidth]{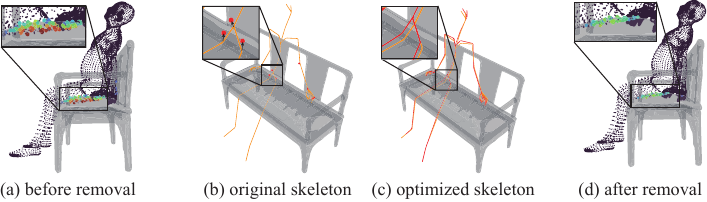}
    \caption{\textbf{Illustration of the penetration removal process.} In panels (a) and (d), the small purple points represent human vertices devoid of penetration, while the larger colored points indicate instances of penetration. The red points symbolize the most pronounced penetration, whereas the blue points signify minimal contact. Panels (b) and (c) feature yellow lines that depict the original skeleton configuration, red markers denoting the target joints undergoing optimization, and red lines illustrating the resultant optimized skeleton configuration.}
    \label{fig:optimize}
\end{figure}

\begin{figure*}[t!]
    \centering
    \includegraphics[width=\linewidth]{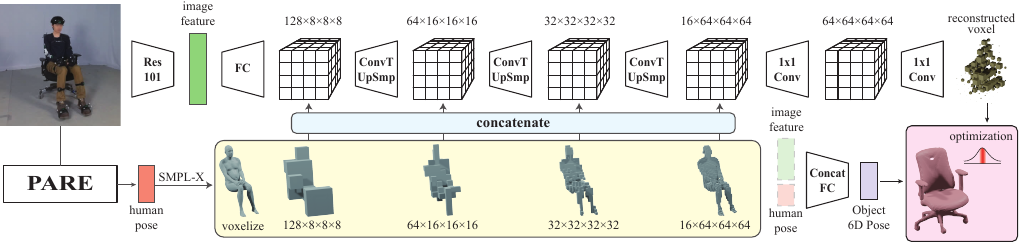}
    \caption{\textbf{Model architecture.} The reconstruction model leverages the predicted voxelized representation of the human to facilitate the estimation of pose for the interacting object. We undertake root 6D pose regression for the object utilizing the image feature in conjunction with SMPL-X parameters. The predictions along with an interaction prior are harnessed for the refinement of the final estimated pose.}
    \label{fig:overview}
\end{figure*}

\subsection{Post-processing}

\paragraph{Data alignment}

Due to the disparate 3D coordinates and temporal clocks of Kinect cameras and the MoCap system, alignment becomes crucial. This alignment is achieved by correlating the 3D coordinates of Kinect sequences with MoCap reconstructions through plane-to-plane correspondences~\cite{segal2009generalized}, a technique that mitigates the influence of outliers, disturbances, and partial overlaps. For the synchronization of temporal sequences from Kinect and MoCap, time-lagged cross-correlation~\cite{shen2015analysis} is applied, a common approach for aligning two sequences with relative time shifts.

\paragraph{Penetration removal}

Owing to the limited sensor count and discrepancies in limb lengths, unrealistic contacts and penetrations persist in captured 3D interactions. To address this issue, we rectify these physical anomalies with a carefully devised optimization algorithm, as depicted in \cref{fig:optimize}. Given a parameterized human body and an articulated object point cloud, we compute penetration depths between the human and object point clouds. Subsequently, we utilize the transpose of the linear-blend-skinning weights of SMPL-X to aggregate the maximum penetration depth and direction to the human skeleton joints. This information is then employed to calculate a target skeleton that mitigates the penetration. Finally, we employ a gradient-based optimization technique to adjust the human model to the new skeleton while maintaining proximity to the MoCap reconstruction. This process reduced the average penetration depth in \ac{dataset} from $3.5$ cm to $2.6$ cm, with an average contact value of $0.2$ cm.

\paragraph{Ensuring data quality}

Following data alignment and penetration removal, the Chamfer distances between annotations and observations in \ac{dataset} are measured at $2.8$ cm for objects and $1.9$ cm for humans. This level of quality is comparable to a recent dataset~\cite{bhatnagar2022behave}, which reports Chamfer distances of $2.4$ cm and $1.8$ cm, respectively.

\paragraph{Privacy protection}

To safeguard identities, we apply face blurring~\cite{blurryfaces} to all participant faces. Furthermore, we informed all participants that they retain the right to have their data removed from \ac{dataset} at any time.

\section{Articulated Object Pose Estimation}

\ac{dataset} offers extensive potential for various \ac{ahoi} tasks, including detection, motion generation, physics-based analysis, and even language-guided motion generation with additional annotations. We highlight the value of \ac{dataset} by focusing on the task of articulated object pose estimation. Despite recent advancements in articulated object pose estimation~\cite{xu2021d3d,fan2022articulated,haresh2022articulated,zhang2023part} and \ac{hoi} reconstruction~\cite{chen2019holistic,taheri2020grab,zhao2022compositional,wu2021saga}, the challenge of articulated object pose estimation in the context of \ac{fahoi} remains largely unaddressed. This specific context demands accurate object pose estimation in scenarios involving heavy occlusion and dense contact.

\subsection{Task Definition}

Given an observed image $I$, the parameterized human model $H=(\beta,\theta_b,\theta_h,R_b,T_b)$, and the meshes $X=\{X_i, i=1,\cdots,N\}$ representing the interacting object with $N$ parts, our task involves estimating the object pose $O=\{(R_i, T_i), i=0,\cdots,N\}$. Here, $\beta\in\mathbb{R}^{10}$, $\theta_b\in\mathbb{R}^{21\times6}$, $\theta_h\in\mathbb{R}^{30\times6}$, $R_b\in\mathbb{R}^6$, and $T_b\in\mathbb{R}^3$ represent the shape and pose parameters of the SMPL-X~\cite{pavlakos2019expressive} model. Specifically, $(R_0\in\mathbb{R}^6, T_0\in\mathbb{R}^3)$ corresponds to the root pose of the object, while $\{(R_i\in\mathbb{R}^6, T_i\in\mathbb{R}^3)\}$ denotes the global rotation and translation for each part $X_i$. The orthogonal 6D representation~\cite{zhou2019continuity} is used for representing rotations in both human and object poses.

\begin{figure}[b!]
    \centering
    \includegraphics[width=\linewidth]{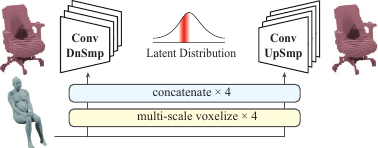}
    \caption{\textbf{Illustration of the interaction prior model.} The interaction prior model, realized as a \ac{cvae}, generates object voxels based on conditioning information from human voxels. During optimization, we aim to minimize the norm of the latent code.}
    \label{fig:prior}
\end{figure}

\subsection{Model Architecture}

Our approach for object pose estimation is rooted in an interaction-aware framework that harnesses the fine-grained geometric relationships present in \acp{hoi}, along with learned interaction priors. This approach comprises two key stages. Given an image and estimated SMPL-X~\cite{pavlakos2019expressive} parameters, we first estimate object occupancy grids and root poses using a reconstruction model. Subsequently, we fine-tune the reconstructed human-object pair using a learned interaction prior. The overall framework of our model is illustrated in \cref{fig:overview}, while \cref{fig:prior} showcases the interaction prior model.

\subsection{Object Reconstruction and Pose Initialization}

Given an observation $I$, we estimate human pose and shape using an off-the-shelf estimator. These estimated human shapes $H'$ are then voxelized into four resolutions with Kaolin~\cite{jatavallabhula2019kaolin}. To better exploit the geometric relationship between human-object pairs, we guide the estimation of object shape and pose using human pose. Specifically, we begin by extracting ResNet-101~\cite{he2016deep} features from the image and subsequently estimate object voxels based on these image features using a 3D decoder. This decoder comprises three 3DConvT layers, along with upsampling layers at distinct resolutions, and two additional 1x1 3DConv layers. Furthermore, we fuse the convolutional feature grids with human voxels at each resolution, amplifying the influence of human pose. The final 3DConv layer generates the estimated object occupancy grid $\mathcal{V}_{O}^{'}$. Additionally, we concatenate image features extracted from ResNet-101 with SMPL-X parameters, and employ an additional MLP to regress the root pose $(R'_0, T'_0)$ of the object. This root pose also serves as the initialization for the optimization process.

To train the reconstruction model, we first initialize the human shape estimator with pre-trained weights from the PARE model~\cite{pavlakos2019expressive}, followed by fine-tuning using the \ac{dataset}. Subsequently, we fix the weights of the PARE model and proceed to train the reconstruction model, utilizing the object pose estimation loss $\LL^\OO$. This loss is characterized by the L1 loss computed on object voxels.

\subsection{Interaction Prior}

To capture the nuanced relationship between humans and interacting objects, we introduce an interaction prior model based on a \ac{cvae}. This model learns the conditional distribution of object occupancy given the human shape.

In this context, the \ac{cvae} prior model conditions on a multi-resolution voxelized human, with the objective of reconstructing a voxelized object. The architecture employs 3DConvNets as both encoder and decoder components. During training, we feed the voxelized object into the encoder to acquire object features at multiple scales. These object features are then combined with the multi-resolution human voxels corresponding to each layer. An MLP estimates the latent Gaussian distribution $\mathcal{N}(\mu,\sigma)$, which is used to parameterize the latent code $z$ through re-parameterization. This latent code is subsequently decoded using the decoder. The feature grids at each decoder layer are concatenated with the corresponding human voxel condition.

Training the prior model occurs on \ac{dataset} and involves four distinct loss components:
\begin{equation}
    \LL_\mathrm{P}=\LL_\mathrm{recon}+\LL_\mathrm{KL}+ \LL_\mathrm{pene}+\LL_\mathrm{contra},
\end{equation}
where $\LL_\mathrm{recon}$ and $\LL_\mathrm{KL}$ denote the standard reconstruction and KL divergence losses, respectively. $\LL_\mathrm{pene}$ constitutes a penetration loss, penalizing voxel grids occupied by both humans and objects. $\LL_\mathrm{contra}$ serves to maximize the distance of latent variables between original and augmented noisy data. Augmentation of training data involves introducing random noise to a portion of the samples.

\subsection{Pose Optimization with Interaction Prior}

\begin{figure}[ht!]
    \centering
    \includegraphics[width=\linewidth]{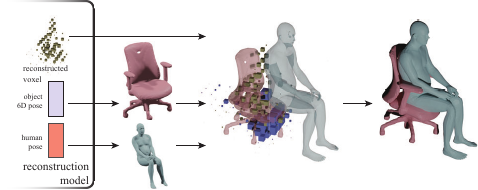}
    \caption{\textbf{An illustration of pose estimation with interaction prior.} Starting with the reconstruction output, we optimize the object according to the \textbf{\textcolor{optyellow}{reconstructed voxel}} and \textbf{\textcolor{optblue}{interaction prior}}.}
    \label{fig:pose_optimize}
\end{figure}

To reconstruct the intricate human-object relationship and refine the object poses, we employ an optimization stage that builds upon initialized poses, utilizing kinematic insights and the interaction prior. This process involves the object's CAD model, \ac{urdf}, estimated SMPL-X parameters $H'$, and object voxels $\mathcal{V}'_{O}$ from the reconstruction model. We initiate the object model $\hat{O}$ using estimated root transformations and random part states. We iteratively update $\hat{O}$'s parameters by minimizing the combined objective $\mathcal{J}_\mathrm{recon} + \mathcal{J}_\mathrm{z}$:
\begin{equation}
    \mathcal{J}_\mathrm{recon} = \Vert V(\hat{O}) - \mathcal{V}'_{O} \Vert_2,\quad{}\mathcal{J}_\mathrm{z} = \Vert\texttt{Enc}(H',\hat{O})\Vert,
\end{equation}
where $V(\cdot)$ is the voxelization function. $\mathcal{J}_\mathrm{recon}$ measures the voxelized object model's distance from the estimated object voxels. $\mathcal{J}_\mathrm{z}$ enforces a small norm for the latent predicted by the \ac{cvae} encoder, regulating proximity to the interaction prior. The process of pose optimization with interaction prior is illustrated in \cref{fig:pose_optimize}.

\paragraph{Object pose optimization}

Optimized parameters include the root 6D pose $R, T$ of the object and joint parameters $\Phi$ (if applicable), controlling part rotation and shift under kinematic constraints. For joints (except the root), we consider revolute, prismatic, or combined revolute-prismatic configurations. The latter, such as the joint linking a chair's base and seat, restricts rotation and shift along the same axis.

During optimization, we initiate the root pose $R, T$ using the estimated root 6D pose from the object pose estimation model. All joint parameters $\Phi$ are set to zero. Optimization involves minimizing the reconstruction loss $\mathcal{J}_\mathrm{recon}$ and interaction prior loss $\mathcal{J}_\mathrm{z}$ through gradient descent. Calculating losses necessitates determining object part occupancy post-application of $R, T$, and $\Phi$ for each optimization step. As direct voxelization lacks differentiability regarding object parameters $R, T$ and $\Phi$, we employ trilinear interpolation on the affined $(0, 1)$ voxel grid to resample voxel occupancy. This permits gradient flow for root and joint parameter updates. Post-optimization, parameters yield an updated 3D object model and enhanced representation (\eg, mesh) with reduced geometric error.

\paragraph{Contrastive loss}

We intend our interac/tion prior to grasp a comprehensive human-object interaction distribution through a conditional Gaussian model. Reasonable and common spatial relationships between human and object latent codes should cluster near the Gaussian mean, in contrast to unreasonable ones. A contrastive loss aids training of the interaction prior model alongside penetration, reconstruction, and KL-divergence losses. Positive examples $(H, \mathcal{V}_{O})$ involve an observed human $H$ and object voxel $\mathcal{V}_{O}$. Corresponding negative examples $(H, \mathcal{V}'_{O})$ are generated by perturbing the object, adding noise to root and articulated poses, and voxelizing $\mathcal{V}'_{O}$. The contrastive loss $\LL_\mathrm{contra}$, defined as $\LL_\mathrm{contra} = max(0, ||\texttt{Enc}(\mathcal{V}_{O}, H)|| - ||\texttt{Enc}(\mathcal{V}'_{O}, H)||)$, guides latent codes of perturbed human-object pairs away from the distribution centroid. Here, $\texttt{Enc}$ represents the conditional encoder of our proposed cVAE-based prior model.

\section{Experiments}

\paragraph{Experimental settings}

We split \ac{dataset} into training, testing, and validation sets; $70\%$ of objects are used for training, $20\%$ for testing, and the remaining for validation. We evaluate our model under two settings: with (\emph{w/ opt}) and without optimization (\emph{w/o opt}). In the \emph{w/ opt}. setting, we report the chamfer distance between objects posed with ground truth and estimated transformation parameters. In the \emph{w/o opt.} setting, we do not have the estimated transformation parameters. Thus, we report the chamfer distance between the ground-truth object mesh and the mesh obtained by running the marching cube algorithm on the reconstructed voxels.

\paragraph{Evaluation metrics}

We evaluate object pose estimation using mean rotation and translation errors for each object part. Object shape reconstruction is evaluated with chamfer distance and intersection over union (IoU). For reconstructed \ac{fahoi}, we assess penetration depth and contact scores between the human and object. Penetration depth is the maximum depth of the object's surface within the human's body, while contact value is the shortest distance between the human and object. Contact values are clipped to [0,20cm] for distant human-object pairs.

\paragraph{Baseline methods}

We compare articulated object pose estimation with LASR~\cite{yang2021lasr} and ANCSH~\cite{li2020category} as baselines, where we use depth maps as input for ANCSH. Both methods are \textit{fine-tuned} on \ac{dataset}. Additionally, we compare our model with D3D-HOI~\cite{xu2021d3d}, PHOSA~\cite{zhang2020perceiving}, and CHORE~\cite{xie2022chore} that jointly estimate human and object poses. We adapted D3D-HOI's optimization objectives to better fit \ac{dataset}'s data distribution.

\begin{figure*}[t!]
    \centering
    \includegraphics[width=\linewidth]{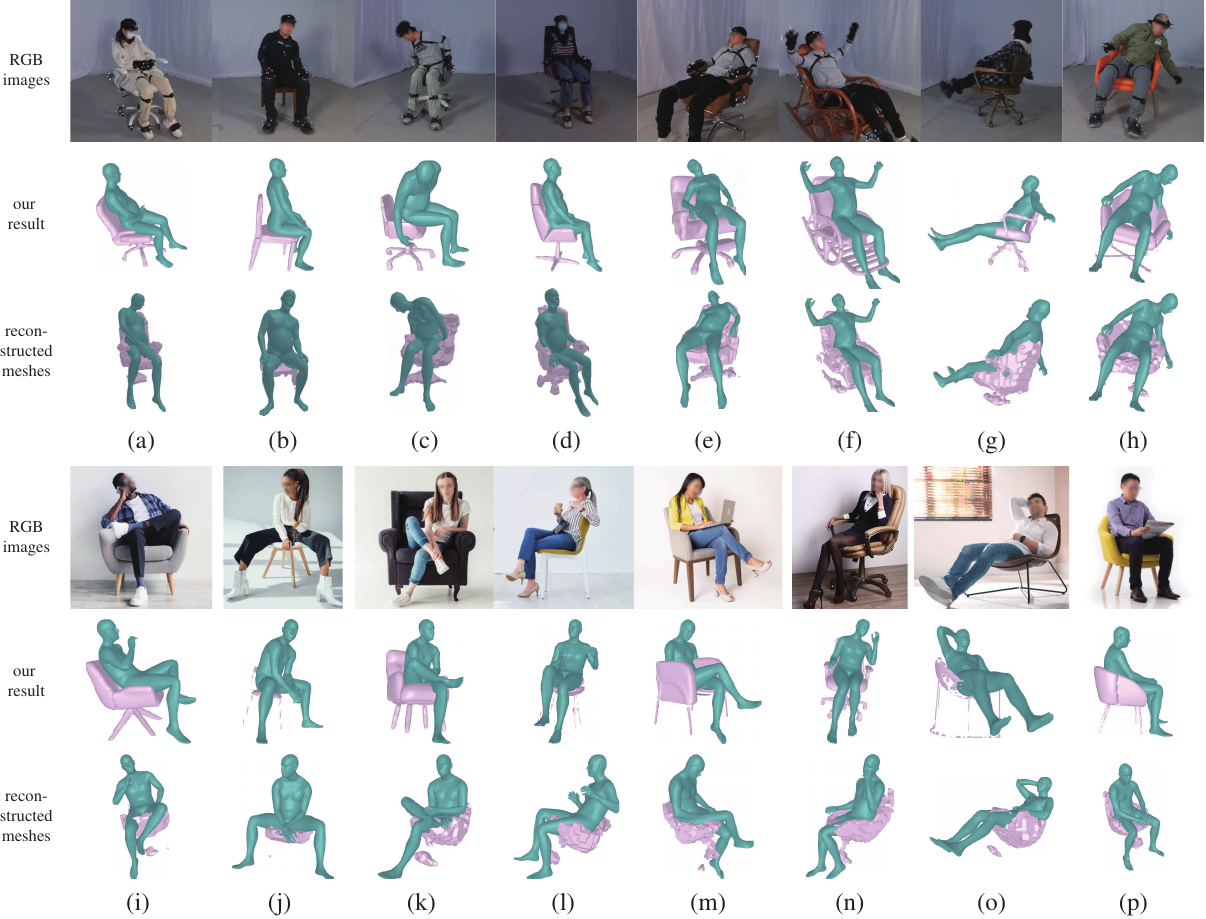}
    \caption{\textbf{Qualitative results.} (a)-(h) Test set results. (i)-(p) Wild images results. RGB images, optimized poses, and mesh obtained by running marching cubes on reconstructed voxels are shown. Please refer to \cref{fig:supp:wild} in Supplementary Materials for more qualitative results.}
    \label{fig:result}
\end{figure*}

\subsection{Results and Analyses}

Quantitative results are presented in \cref{tab:result}. Our model, leveraging geometrical relationships, exhibits substantial improvements in pose estimation and shape reconstruction compared to existing methods. In the \emph{w/o opt.} setting where the object is unknown, our model surpasses the state-of-the-art LASR method by a significant margin. While D3D-HOI and ANCSH excel, they assume known object structures. Remarkably, our model outperforms all baselines when provided with the object structure in the \emph{w/ opt.} setting.

\begin{table}[hb!]
    \centering
    \small
    \caption{\textbf{Comparisons against existing methods.} $*$: method requires knowledge of object structure and/or geometry; $\dagger$: method does not rely on object-related knowledge.}
    \resizebox{\linewidth}{!}{%
        \begin{tabular}{lxxxxyy}
            \toprule
            \multirow{2}{*}{Method}&\multicolumn{4}{x}{Object} & \multicolumn{2}{y}{HOI}\\
            & \specialcell{Rot.$\downarrow$\\($^\circ$)} & \specialcell{Transl.$\downarrow$\\(mm)} &
            \specialcell{CD$\downarrow$\\(mm)} & \specialcell{IoU$\uparrow$\\(\%)} & 
            \specialcell{Pene.$\downarrow$\\(mm)} & \specialcell{Cont.$\downarrow$\\(mm)}\\
            \midrule
            LASR$^\dagger$~\cite{yang2021lasr} & /&/&205.2&/&/&/ \\
            Ours (w/o opt.)$^\dagger$& / & / & \textbf{160.2} & \textbf{11.03} & \textbf{4.530} & \textbf{2.720} \\
            \midrule
            ANCSH$^*$~\cite{li2020category} & /&/&90.36&/&/&/ \\
            PHOSA$^*$~\cite{zhang2020perceiving} & 29.31 & 175.2 & 177.9 & 7.60 & 2.046 & 1.689 \\
            D3D-HOI$^*$~\cite{xu2021d3d}  & 27.31 & 119.2 & 126.9 & 16.60 & 7.472 & \textbf{1.163} \\
            CHORE$^*$~\cite{xie2022chore} & 21.82 & 87.58 & 95.40 & 16.44 & \textbf{1.050} & 1.742 \\
            Ours (w/ opt.)$^*$& \textbf{19.35} & \textbf{66.23} & \textbf{72.30} & \textbf{21.57} & 1.143  & 1.562\\
            \bottomrule
        \end{tabular}%
    }%
    \label{tab:result}
\end{table}

We present qualitative results in \cref{fig:result}, where columns (a)-(h) illustrate the reconstruction outcomes on the test set. In these columns, we display the reconstructed meshes prior to optimization using the marching cubes algorithm. It is evident from the visualizations that our model successfully produces plausible and accurate interaction representations even before the optimization process. Notably, the optimization step enhances the finer interaction details.

\subsection{Ablations}

We conduct three ablation studies to assess the efficacy of our model's design choices. Quantitative results of these ablation studies are presented in \cref{tab:ablation}.

\begin{table}[hb!]
    \centering
    \small
    \caption{\textbf{Ablation of interaction, prior, and contrastive loss.}}
    \label{tab:ablation}
    \resizebox{\linewidth}{!}{
        \begin{tabular}{lxxxxyy}
            \toprule
            \multirow{2}{*}{Method}&\multicolumn{4}{x}{Object} & \multicolumn{2}{y}{HOI}\\
            &
            \specialcell{Rot.$\downarrow$\\($^\circ$)} & \specialcell{Transl.$\downarrow$\\(mm)} &
            \specialcell{CD$\downarrow$\\(mm)} & \specialcell{IoU$\uparrow$\\(\%)} & 
            \specialcell{Pene.$\downarrow$\\(mm)} & \specialcell{Cont.$\downarrow$\\(mm)}\\
            \midrule
            Full$^\dagger$         & / & / & \textbf{160.2} & \textbf{11.03} & 4.530 & \textbf{2.720}\\
            $-\,$prior$^\dagger$   & / & / & 165.3 & 10.52 & \textbf{4.377} & 3.295\\
            \midrule
            Full$^*$         & 19.35 & \textbf{66.23} & \textbf{72.30} & \textbf{21.57} & 1.143 & \textbf{1.562} \\ 
            $-\,$prior$^*$         & 19.97 & 83.39 & 87.90 & 18.81 & 1.749 & 2.081 \\
            $-\,$contr.$^*$      & 21.52 & 81.90 & 87.28 & 18.93  & 1.265 & 2.393  \\
            $-\,$inter.$^*$   & \textbf{17.88} & 69.53 & 78.12 & 19.50 & \textbf{1.022} & 2.320\\
            \bottomrule
        \end{tabular}%
    }%
\end{table}

\paragraph{Prior}

We conduct an experiment where we remove the interaction prior model and solely optimize object poses by minimizing $\LL_{\mathrm{recon}}$. In both $*$ and $\dagger$ settings, we observe a substantial performance drop. This underscores the critical role played by the interaction prior in accurately estimating object poses. It is important to note that both settings involve an optimization step, and the primary distinction is that the $*$ model possesses access to the object's geometry and structure during optimization. When the prior model is omitted in the $\dagger$ setting, we observe a decrease in penetration and a larger increase in contact value. This observation suggests that our interaction prior model exerts influence by pulling the object closer to the human when they are not in contact.

\paragraph{Contrast}

In this ablation, we exclude the contrastive loss $\LL_{\mathrm{contra}}$ from the training of the prior model. The results are analogous to those of the $-$prior experiment. This outcome underscores the crucial role that the contrastive loss plays in facilitating the learning of a robust interaction prior.

\paragraph{Interaction}

We proceed to remove the concatenation of human voxels in the 3DConv layers of both the reconstruction model and the interaction prior model. This removal eliminates the interaction awareness in our model. We observe a modest degradation across all object reconstruction metrics, underscoring the importance of interaction awareness in our approach. Interestingly, the removal of interaction awareness leads to increased contact values and decreased penetration, resembling the outcomes of the $-$prior ablation in the \textit{w/o opt.} setting. This suggests that interaction awareness also contributes to bringing the human and object into closer proximity. Lastly, we note an unexpected low rotation error, which we attribute to the presence of rotation symmetries in the dataset.

Additionally, we assess our method's performance under varying qualities of human pose estimation in \cref{tab:cmphuman}. The results reveal notable improvement in object pose estimation as human poses become more accurate, thus validating our initial hypothesis. Notably, the pose estimation model \cite{kocabas2021pare} effectively predicts most challenging poses accurately, leaving the avenue of leveraging interactions to enhance human pose estimation as a potential future research direction.

\begin{table}[htb!]
    \centering
    \caption{\textbf{Ablation of human pose estimation quality.} GT denotes using ground-truth human poses to optimize the object poses, No prior denotes not considering human-object interaction prior.}
    \resizebox{\linewidth}{!}{%
        \begin{tabular}{lxxyy}
            \toprule
            \multirow{2}{*}{Method}&\multicolumn{2}{x}{Human} & \multicolumn{2}{y}{Object}\\
            &
            MPJPE$\downarrow$(mm) & PA-MPJPE$\downarrow$(mm) & 
            CD$\downarrow$(mm) & IOU$\uparrow$(\%)\\
            \midrule
            No prior &/&/&87.90&18.81 \\
            PARE~\cite{kocabas2021pare} &81.09&47.19&73.79&21.66 \\
            PARE(finetune) &74.50&43.99&72.30&21.57 \\
            GT             &\textbf{0}&\textbf{0}&\textbf{65.50}&\textbf{23.16} \\
            \bottomrule
        \end{tabular}%
    }%
    \label{tab:cmphuman}
\end{table}

In summary, our analysis highlights the substantial contributions of all three model components to object pose and shape reconstruction.

\paragraph{In-the-wild generalization}

We also assess the model's generalizability on a limited set of internet images. As depicted in \cref{fig:result} (i-p), the qualitative results illustrate that our model successfully generalizes its capabilities to images captured outside of controlled laboratory settings.

\paragraph{Failure cases}

\begin{figure}[t!]
    \centering
    \includegraphics[width=\linewidth]{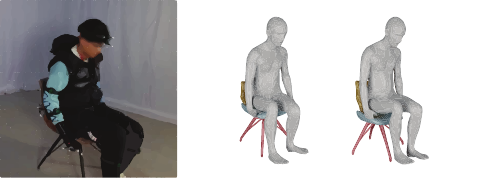}
    \caption{\textbf{Failure cases.} In situations involving rotation-symmetrical objects, our model encounters challenges with estimating rotation while maintaining a relatively low visual error.}
    \label{fig:failure}
\end{figure}

Our model encounters challenges in accurately estimating the orientation of object parts when those parts exhibit geometric similarity under specific rotations. Rotation symmetry is commonly observed in spherical and cylindrical object components, such as the base of a stool or a round seat. An illustrative example of this symmetry is presented in \cref{fig:failure}. Notably, existing methods~\cite{fan2017point,wang2019normalized} address this challenge through (i) accepting multiple equally-valid ground truths and (ii) employing a min-of-N loss to calculate the smallest distance to any of these ground truths. However, implementing such methods necessitates a meticulous classification of symmetry types for each object.

Furthermore, we observe that our model's performance diminishes in scenarios where no \ac{fahoi} is present, such as instances where a human is situated far away from an object like a chair. Under these circumstances, our model is unable to leverage interactions to enhance object pose estimation.

\section{Application: Generating Interacting Humans}

\begin{figure}[ht!]
    \centering
    \begin{subfigure}{0.485\linewidth}
        \includegraphics[width=\linewidth]{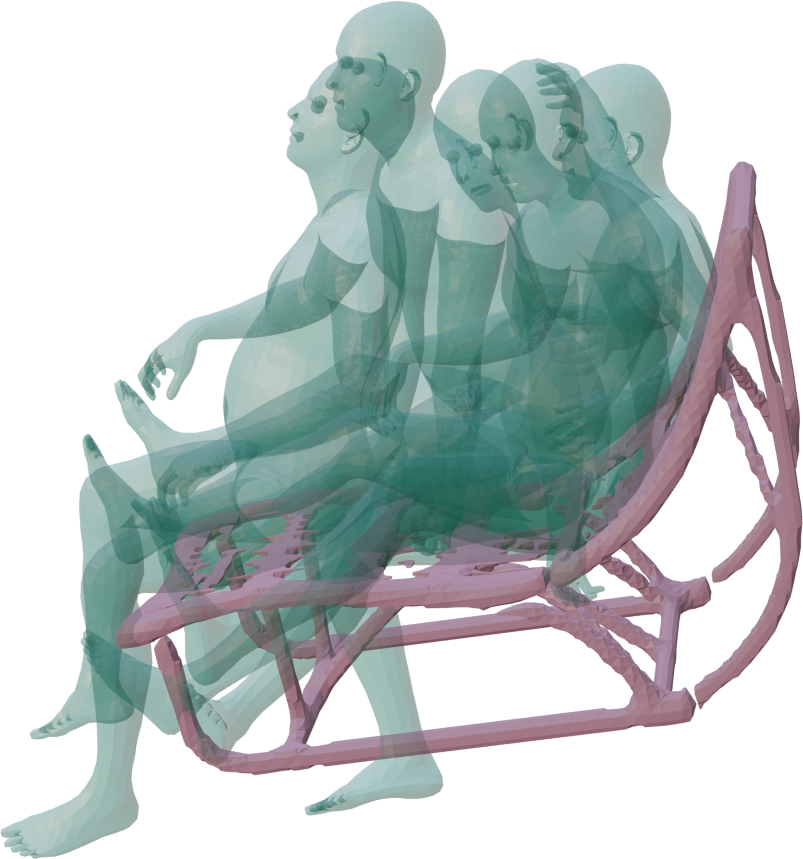}
    \end{subfigure}%
    \begin{subfigure}{0.50\linewidth}
        \includegraphics[width=\linewidth]{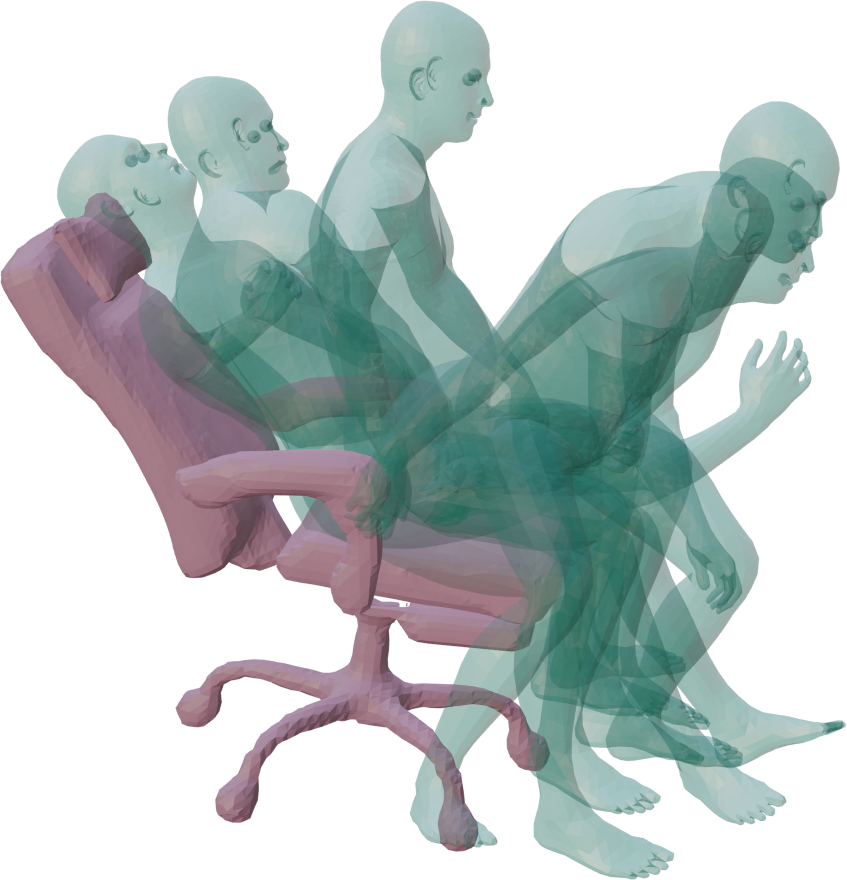}
    \end{subfigure}%
    \\%
    \begin{subfigure}{0.485\linewidth}
        \includegraphics[width=\linewidth]{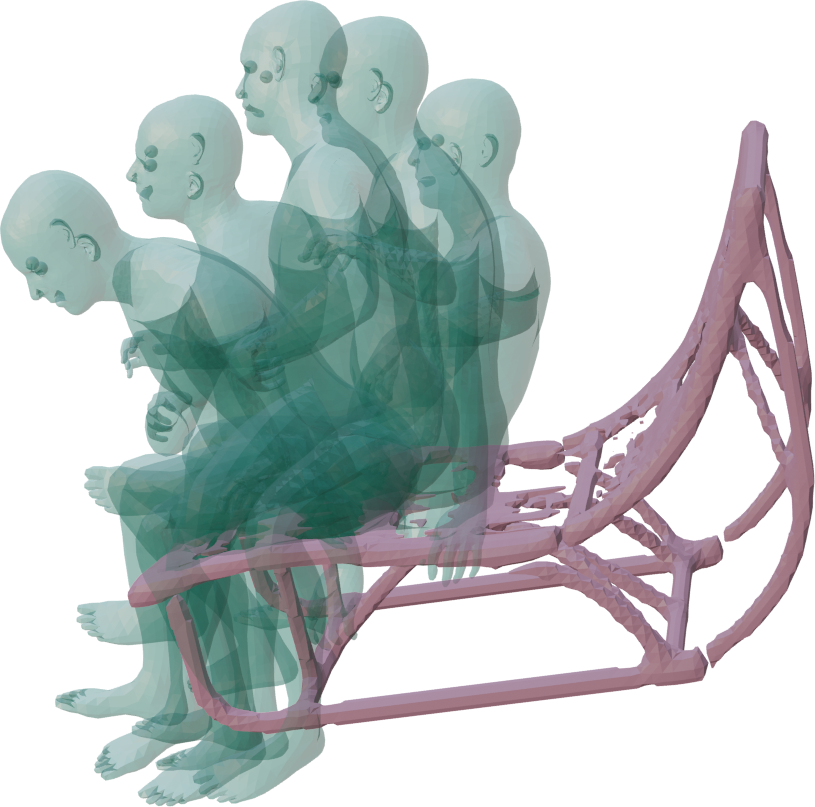}
    \end{subfigure}%
    \begin{subfigure}{0.49\linewidth}
        \includegraphics[width=\linewidth]{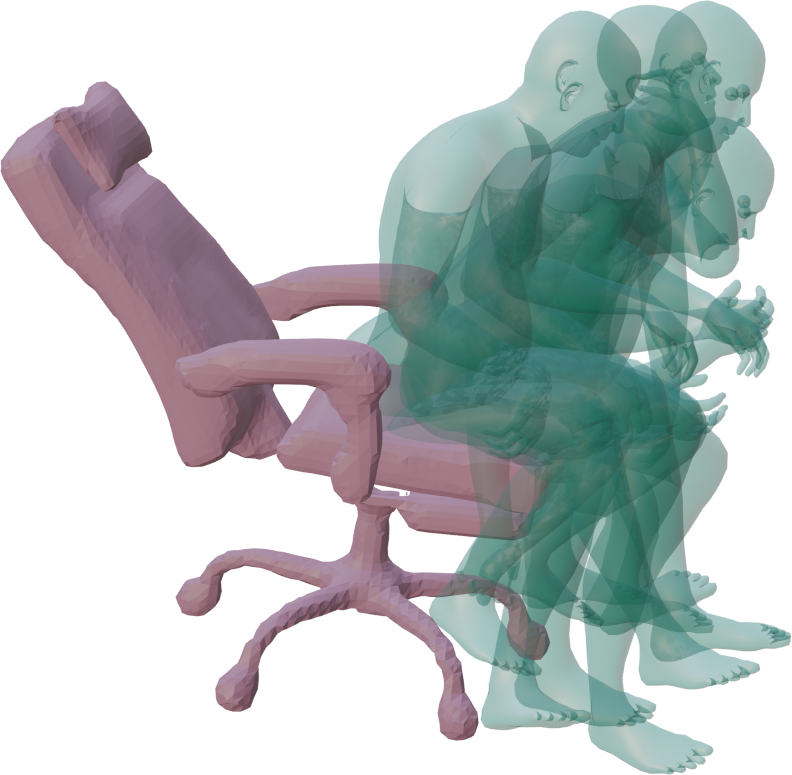}
    \end{subfigure}%
    \caption{\textbf{Generated human poses given articulated objects.} The models are trained on \ac{dataset} \textbf{(a, b)} and COUCH~\cite{zhang2022couch} \textbf{(c, d)}.}
    \label{fig:generate}
\end{figure}

We further investigate the intricate relationships within \ac{ahoi} by exploring the generation of interacting human poses in the presence of articulated objects. To this end, we employ a 3D conditional diffusion model known as SceneDiffuser~\cite{huang2023diffusion}, trained on our \ac{dataset}. To evaluate the quality of the generated poses, we compare them with poses generated using the same model trained on COUCH~\cite{zhang2022couch}, a recent dataset featuring humans seated on \textit{rigid} chairs. We use the feature extracted from the point cloud of the object via Pointnet++ as a conditioning input and flatten the SMPL-X parameters of the human to form tokens for input to a Transformer. The implementation details closely follow those of the human pose generation task described in Huang \etal~\cite{huang2023diffusion}. Qualitative comparisons of the generated human poses are shown in \cref{fig:generate}. Notably, the model trained with \ac{dataset} captures more nuanced and natural geometrical relationships when interacting with articulated objects. For a more in-depth analysis of this downstream application, we direct readers to the \cref{supp:additional} in Supplementary Materials.

\section{Conclusion}

We advance the study of \ac{hoi} to encompass fine-grained, articulated interactions with (i) \ac{dataset}, an extensive dataset, (ii) a challenging object reconstruction problem under \ac{fahoi}, and (iii) a strong baseline. Our \ac{dataset} captures diverse, natural \acp{ahoi} involving various sittable objects. The object reconstruction problem confronts kinematic assumptions, with our model effectively leveraging intricate interactions to resolve ambiguities.

\paragraph{Limitations}

One limitation of our work lies in the fact that the parametric human model used in \ac{dataset} does not account for clothing, leading to misalignments between the 3D annotations and the images. Consequently, the usage of pixel-aligned features may be compromised.

\paragraph{Acknowledgment}
The authors would like to thank four anonymous reviews for constructive feedback and NVIDIA for their generous support of GPUs and hardware. This work is supported in part by the National Key R\&D Program of China (2022ZD0114900) and the Beijing Nova Program.

{
    \small
    \bibliographystyle{config/ieee_fullname}
    \bibliography{reference_header,reference}
}

\clearpage
\appendix
\renewcommand\thefigure{A\arabic{figure}}
\setcounter{figure}{0}
\renewcommand\thetable{A\arabic{table}}
\setcounter{table}{0}
\renewcommand\theequation{A\arabic{equation}}
\setcounter{equation}{0}
\pagenumbering{arabic}
\renewcommand*{\thepage}{\arabic{page}}
\setcounter{footnote}{0}

\section{Model}

\paragraph{Why do we mainly focus on the articulated object pose estimation?}

Existing studies of \ac{hoi} usually estimate the pose of humans and objects jointly, hoping the two estimations to improve each other. However, due to the imbalanced attention received by the human and articulated object pose estimation, we empirically observe that the object pose estimation is far from well-solved compared with human pose estimation, especially in scenarios where dense interactions and occlusions appear. Therefore, we mainly focus on improving the untouched articulated object pose estimation under human pose guidance in this paper, leveraging the mature and stable techniques of human pose estimation. Such motivation is similar to Ye \etal~\cite{ye2022ihoi}, which focuses on improving the reconstruction of interacting objects rather than the hand. Of note, our dataset still supports human pose estimation and encourages efforts that potentially improve it. Tab. 4 from the main text shows that incorporating the human pose information can significantly improve the object pose estimation performance, which verifies our assumption. The ground-truth human pose can further improve the object pose estimation by a large margin, demonstrating that further optimization of human poses is promising. It is regarded as one important step in our future work.

\paragraph{Coordinates for reconstruction and optimization}

Both object reconstruction and optimization are conducted in the human local coordinate centered at the pelvis bone of the SMPL model with the same orientation as the human root. We set a $2m \times 2m \times 2m$ cubic as the boundary for voxelization and interaction prior.

 \paragraph{Details of Reconstruction Model}The ResNet-101 to extract features from the input image outputs feature vectors of 1024 dimension. The feature vector is then mapped to $128\times8\times8\times8$ to form the input of the 3D blocks. Each 3D block consists of a 3D convolutional layer with a kernel size 3x3x3, a max-pooling layer and a batchnorm layer. The number of channels of the first 3D convolutional layer is 129, inlucding 128 for the object and 1 for the concatenated human occupancy. The chanels for the object are reduced by half after each block. The max-pooling layer has a pooling size of 2x2x2 and a stride of 2 in all dimensions. 

\paragraph{Adapting D3D-HOI as baseline for our task}

The D3D-HOI method \cite{xu2021d3d} is originally designed for hand-centric interactions, such as opening and closing a microwave, and contains manually defined optimization objectives, such as distance between hand and object. 
We make the following modifications to D3D-HOI to better fit the context of \ac{dataset}: 
\begin{enumerate}[leftmargin=*,noitemsep,nolistsep]
    \item We replace the differentiable articulated object model in D3D-HOI by the pytorch\_kinematics package (\url{https://github.com/UM-ARM-Lab/pytorch\_kinematics}), which
    supports articulated objects with multiple links and joints.
    \item We changed the contact error in D3D-HOI to the distance between the hip joint and the center of the chair seat. Since the hip joint is usually higher than its nearby skin, we compute this error by adding a $20cm$ offset along the negative Y direction. 
    \item The orientation term in D3D-HOI encourages the human and the object to have opposite directions in ``opening'' and ``closing'' actions. We change this term to encourage the human to have the same orientation as the chair in the ``sitting'' case.
\end{enumerate}

\paragraph{Data preparation for CHORE and PHOSA}

Since PHOSA requires predefined contact pairs as heuristics to reconstruct human-object interaction, we manually labeled each object mesh with contact maps corresponding to human body parts during the interaction. A part of the labeling results are shown in \cref{fig:supp:phosa-contact}.

\section{Additional Results} \label{supp:additional}

\paragraph{Generative model} We evaluate the value of \ac{ahoi} in \ac{dataset} by training conditional generative models~\cite{huang2023diffusion} on both the \ac{dataset} dataset and the COUCH~\cite{zhang2022couch} dataset. \Cref{fig:supp:generative_model} shows that both models can generate realistic interactions with objects, and the model trained with \ac{dataset} can generate interacting poses with more full-body interactions. This observation confirms the value and the contribution of our \ac{dataset} dataset. 

\paragraph{Qualitative results}

In \cref{fig:supp:result}, we qualitatively show more randomly selected results on the test set of \ac{dataset}. In general, our model predicts accurate object poses and shapes.

\paragraph{Qualitative comparisons} We further compare reconstructions of our method against results from CHORE~\cite{xie2022chore} and PHOSA~\cite{zhang2020perceiving} in \cref{fig:supp:qual-comp}. The qualitative comparison shows that our method can reconstruct interactions accurately.

\paragraph{In the wild}

\begin{table}[t!]
    \centering
    \caption{\textbf{Object reconstruction errors on BEHAVE dataset, with object kinematic structure and optimization.}}
    \resizebox{\linewidth}{!}{%
        \begin{tabular}{lxxyyxxyy}
            \toprule
            \multirow{2}{*}&\multicolumn{2}{x}{Chair} &\multicolumn{2}{y}{Table} &\multicolumn{2}{x}{Yogaball} & \multicolumn{2}{y}{Suitcase}\\
            &
            \specialcell{CD$\downarrow$\\(mm)} & \specialcell{IOU$\uparrow$\\(\%)} & 
            \specialcell{CD.$\downarrow$\\(mm)} & \specialcell{IOU.$\uparrow$\\(\%)} &
            \specialcell{CD$\downarrow$\\(mm)} & \specialcell{IOU$\uparrow$\\(\%)} & 
            \specialcell{CD.$\downarrow$\\(mm)} & \specialcell{IOU.$\uparrow$\\(\%)}\\
            \midrule
            w/o HOI prior &134.5&11.35&161.6&10.53&106.37&30.53&161.0&\textbf{29.80} \\
            w/ HOI prior &\textbf{127.3}&\textbf{14.22}&\textbf{152.2}&\textbf{12.86}&\textbf{98.79}&\textbf{33.75}&\textbf{158.4}&29.62\\
            \bottomrule
        \end{tabular}%
    }%
    \label{tab:behave}
\end{table}

In \cref{fig:supp:wild}, we qualitatively evaluate the generalization power of our model with internet images and images captured in the wild.

\paragraph{Experimental results on the BEHAVE dataset}

We apply our method to the BEHAVE dataset~\cite{bhatnagar2022behave} to evaluate the generalizability of the reconstruction and HOI prior model. We select four objects from the object list with rich full-body HOI, namely a chair, a square table, a yoga ball, and a suitcase. Our method is tested under the full object knowledge setting. We separately train object reconstruction and HOI prior models for each object. Different kinds of interaction (\eg, move and sit for the square table) are mixed up in one model. We show quantitative results in \cref{tab:behave} and qualitative results in \cref{fig:supp:behave}. We observe that although the metrics drop numerically, our model can still reconstruct the poses of the interacting objects.

\section{Dataset}

\subsection{Data collection}

\paragraph{Object gallery}

We render all objects in \ac{dataset} in \cref{fig:supp:chairs}. Parts are colored according to category. 

\paragraph{Instructions} \label{supp:instructions}

Each participant was instructed to sit down before and after each instruction for synchronization. Participants can stand up and walk around while performing an instruction. All physical interactions were performed with the sittable objects. All other objects that appeared in the instructions (table, person, phone, \etc) required participants to interact by imaging their presence.
\begin{enumerate}[leftmargin=*,noitemsep,nolistsep]
    \item Pick up an object from the ground.
    \item Talk to someone next to you.
    \item Relax alone at home.
    \item Listen to your friend talk while propping your head with your hand.
    \item Sit and play with your phone.
    \item Sit with your hands on the seat.
    \item Think with your head lowered.
    \item Your neck feels uncomfortable.
    \item Grab a thing from the desk behind you. 
    \item Move the chair forward.
    \item Lean on the back. Adjust or rock it if you can.
    \item Move the chair.
    \item Adjust the chair.
    \item Sit with a twisted posture.
    \item Sit with your feet on the footstep or the footrest.
    \item Change the pose of your legs.
    \item Stretch a little in the chair.
    \item Change to another pose of sitting.
    \item Adjust the height of the seat.
    \item Walk around the chair and sit down.
    \item Move, rock, or rotate the chair.
    \item Your back feels uncomfortable.
    \item Lean your head on the headrest. Adjust it if possible.
    \item Stretch your back in the chair.
    \item Talk to the person behind you.
    \item Move the chair backward.
    \item Lay in the chair.
    \item Put your arms on the armrests. Adjust them if you can.
    \item Move the chair to your left.
    \item Move the chair to your right.
    \item Adjust the seat.
    \item Pick up a heavy object from the ground.
\end{enumerate}

We only sample instructions that are \textit{compatible} given an object. For example, ``Lean on the back'' is \textit{not compatible} for all stools. \Cref{fig:supp:trajectories} shows diverse performances in \ac{dataset}.

\paragraph{Recruitment} 

Due to the complex nature of data collection that requires physical presence at the scene while wearing MoCap suits, all participants were voluntary colleagues. Participants were compensated with a gift with a value of \$4 USD for every 18 sequences recorded. 

\paragraph{Body and hand shape}

We use optical trackers to record the positions of each participant's head, two hands, and two feet. We then optimize the body shape parameter $\beta$ of the SMPLX model to fit the tracker positions. We rely on SMPLX's default hand shape parameter since our primary focus is not to model dexterous hand-object interactions.

\paragraph{Motion capture system} 

We used a Noitom Virtual Production Solution (VPS) camera system and a Noitom Perception Neuron Studio IMU system. The cameras each have 1280x1024 resolution, 210 fps, $<$5ms latency, 3.6mm F\#2.4 lens, 81 deg horizontal FoV, and 67 deg vertical FoV.

\subsection{Post Processing}

\paragraph{Spatial alignment}

Our data collection system consists of multiple pieces of hardware, including 4 Azure Kinect DK cameras and a hybrid MoCap system. Each camera and the MoCap system have their own coordinate systems. We use OpenCV and an Aruco checkerboard to register all cameras to the camera space of the left-most camera and align it with the MoCap's coordinate frame with an \ac{icp} algorithm.

Given the transformation matrices of the Kinect cameras, we apply a custom \ac{icp} algorithm to refine both the multi-view point clouds and the registration of Kinect and Mocap. We base our method on plane-to-plane correspondences \cite{segal2009generalized} to alleviate the sensitivity to outliers, disturbances, and partial overlaps. Given the source point set $\textit{P}=\{p_i, i=1,...,N\}$ captured by the Kinect depth cameras and the target set $\textit{Q}=\{q_i, i=1,...,M\}$ reconstructed from the MoCap system, the goal is to calculate the optimum transformation matrix $T$, such that $TP^T=Q^T$. Following point-to-point \ac{icp} \cite{chen1992object}, we first find the nearest points $\widetilde{q_i}$ in $Q$ to each $p_i$ in $P$. Next, we iteratively update $T$ to minimize the Mahalanobis distance between $P$ and $Q$:
\begin{equation}
    {T} = \arg \mathop {\min }\limits_T \sum\limits_{i = 1}^M {d_i^T{(C_{n,\tilde i}^Q + TC_{n,i}^P{T^T})^{ - 1}}{d_i}}
\end{equation}
where \textit{$d_i$} is the corresponding Euclidean distance between $p_i$ and $\widetilde{q_i}$, $C_{n,\tilde i}^Q$ and $C_{n,i}^P$ the covariance matrix calculate by the $n$ nearest points around $\widetilde{q_i}$ in $Q$ and $p_i$ in $P$. 
Finally, we use Anderson Acceleration \cite{fang2009two} for a faster convergence to a fixed point.

\paragraph{Temporal alignment}

Observed images and poses in \ac{dataset} come from two independent systems (\ie, MoCap and Kinect) without clock synchronization. Since both systems run steadily at 30 Hz, the two recorded data streams have a constant difference in time. We use a \ac{tlcc} \cite{shen2015analysis} algorithm to align the two systems temporally.

Specifically, we first extract the heights of the subject's head and two hands from both systems. For our MoCap system, we can directly read the joint positions with forward kinematics. For the Kinect cameras, we obtain the human joint positions with the Kinect Body Tracker SDK. Next, we compute the first-order differential on each sequence and compute the time offset between the differentials of each joint using \ac{tlcc}. Finally, by measuring the peak of the \ac{tlcc} correlation, we obtain three offsets (one for each joint); we use the median of the three offsets as our final temporal offset. 

\section{Compliance}

\paragraph{List of code, data, models used, and their licenses}

We used the following assets. Please find the licenses of corresponding assets in the directories inside square brackets.
\begin{itemize}[leftmargin=*,noitemsep,nolistsep]
    \item SMPL-X \cite{pavlakos2019expressive} model and body [license/smplx-model,license/smplx-body.txt]
    \item ExPose \cite{choutas2020monocular} model and code [license/expose.txt]
    \item FrankMocap \cite{rong2021frankmocap} model and code [license/frankmocap.txt]
    \item PARE \cite{kocabas2021pare} model and code [license/pare.txt]
    \item Category-Level Articulated Object Pose Estimation \cite{li2020category} model and code [\textit{No license information found.}]
    \item Metropoly rigged 3D people (used in main paper Fig.3 and supplementary video) 
    \item D3D-HOI \cite{xu2021d3d} code [\textit{No license information found.}]
    \item iStock [\url{https://www.istockphoto.com}] images used for in-the-wild evaluations. [license/istock.txt]
\end{itemize}

\clearpage

\begin{figure*}[t!]
    \centering
    \includegraphics[width=\linewidth]{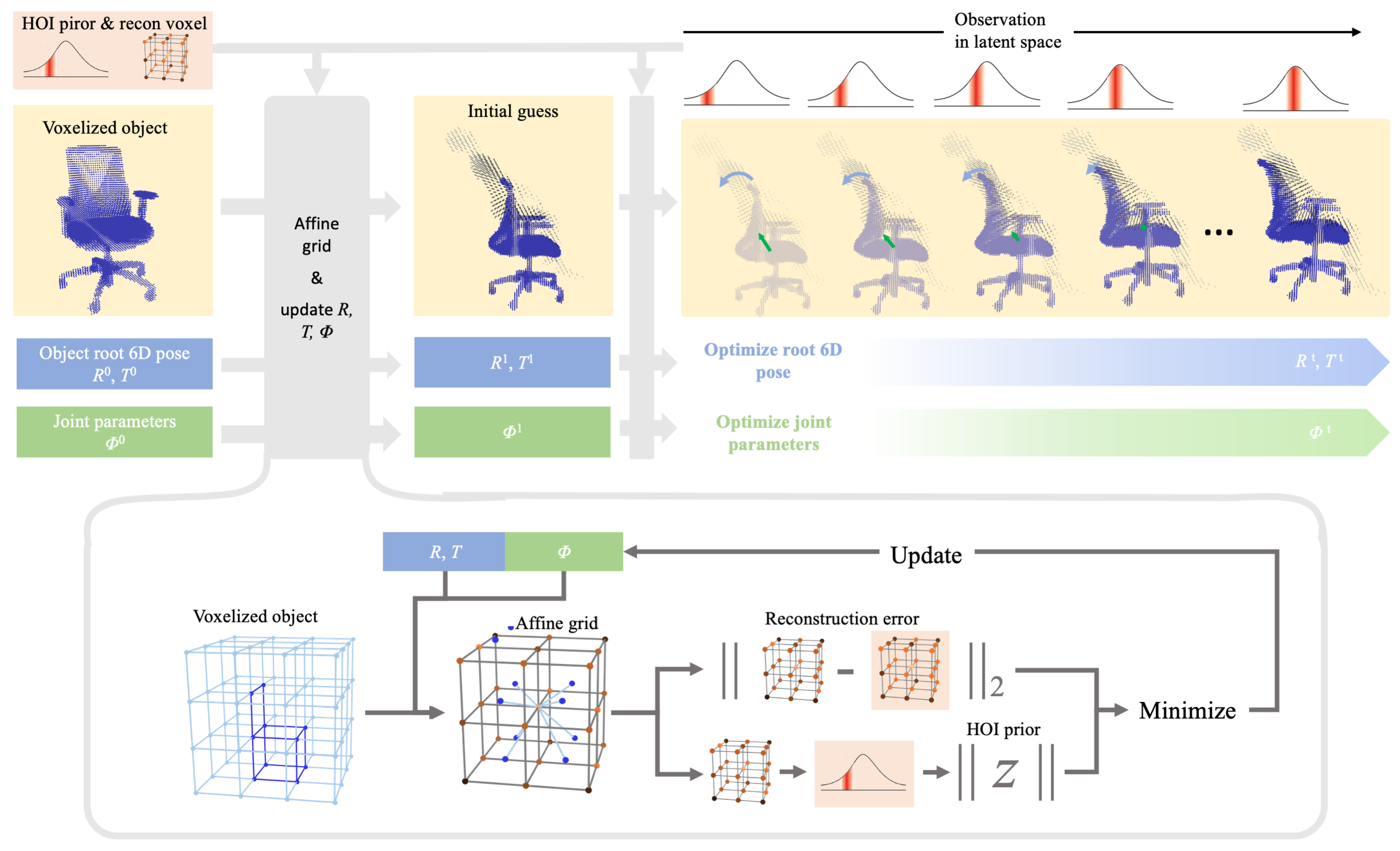}
    \caption{\textbf{Detailed diagram of the optimization process.}}
    \label{fig:supp:overview}
\end{figure*}

\clearpage

\begin{figure*}[t!]
    \centering
    \includegraphics[width=\linewidth]{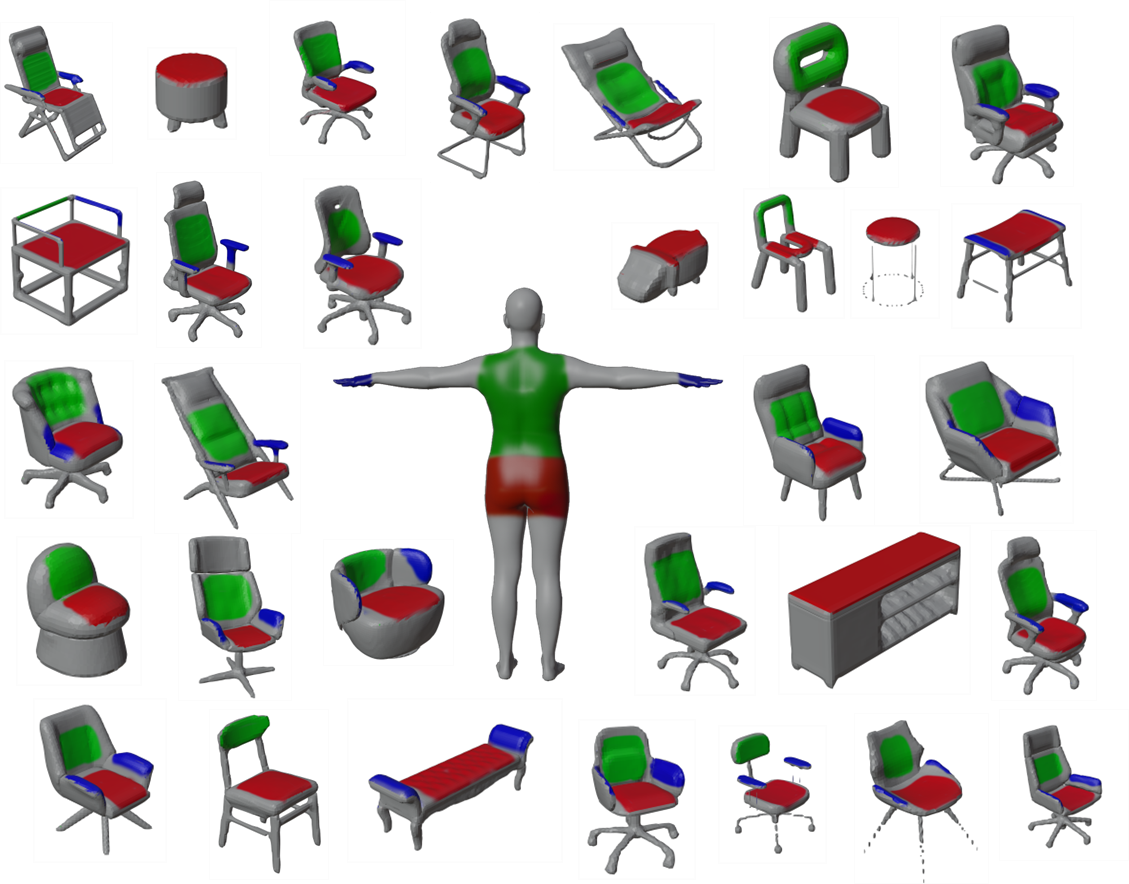}
    \caption{\textbf{Labeled contact maps} We use three colors to show the mappings of the surfaces on human bodies and objects that frequently get in touch during interactions.}
    \label{fig:supp:phosa-contact}
\end{figure*}

\clearpage

\begin{figure*}[t!]
    \centering
    \begin{subfigure}{0.4\linewidth}
        \includegraphics[width=\linewidth]{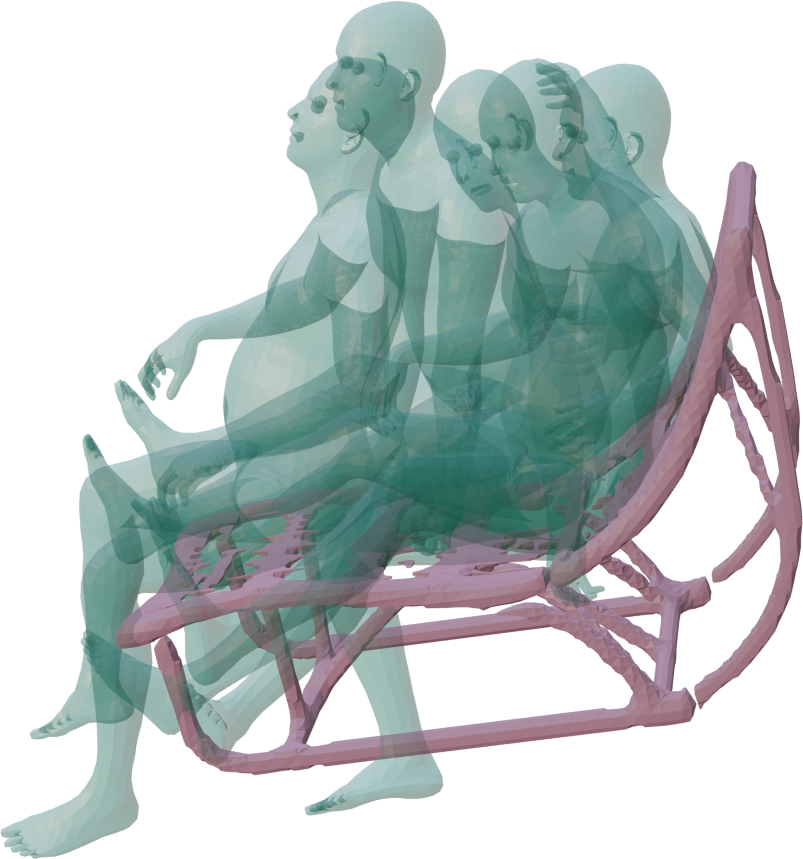}
    \end{subfigure}
    \begin{subfigure}{0.4\linewidth}
        \includegraphics[width=\linewidth]{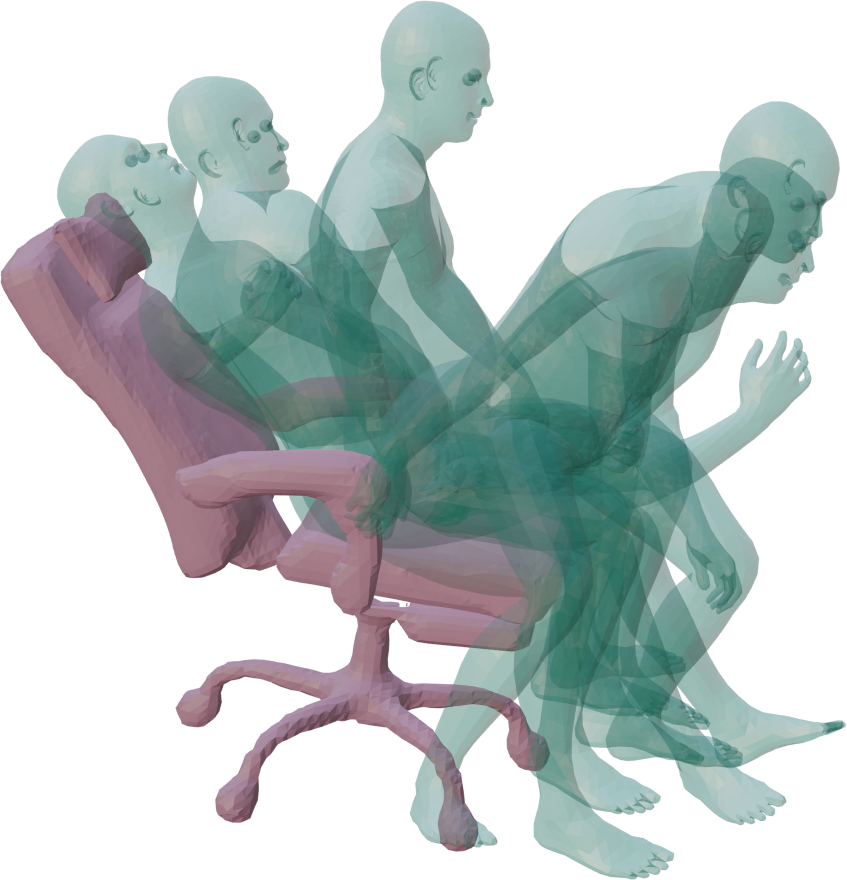}
    \end{subfigure}
    \\
    \begin{subfigure}{0.4\linewidth}
        \includegraphics[width=\linewidth]{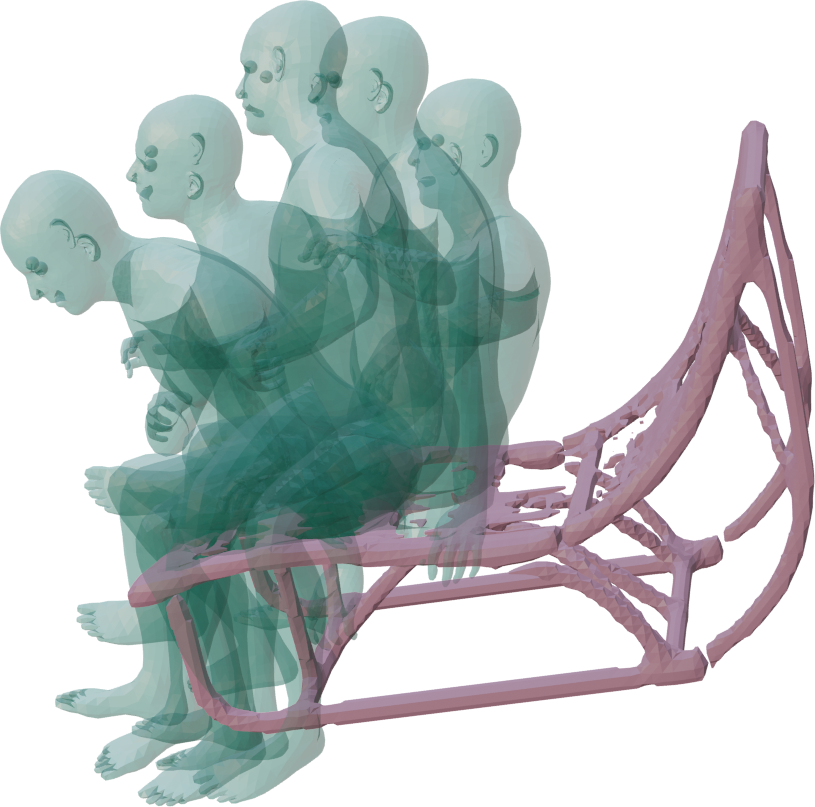}
    \end{subfigure}
    \begin{subfigure}{0.4\linewidth}
        \includegraphics[width=\linewidth]{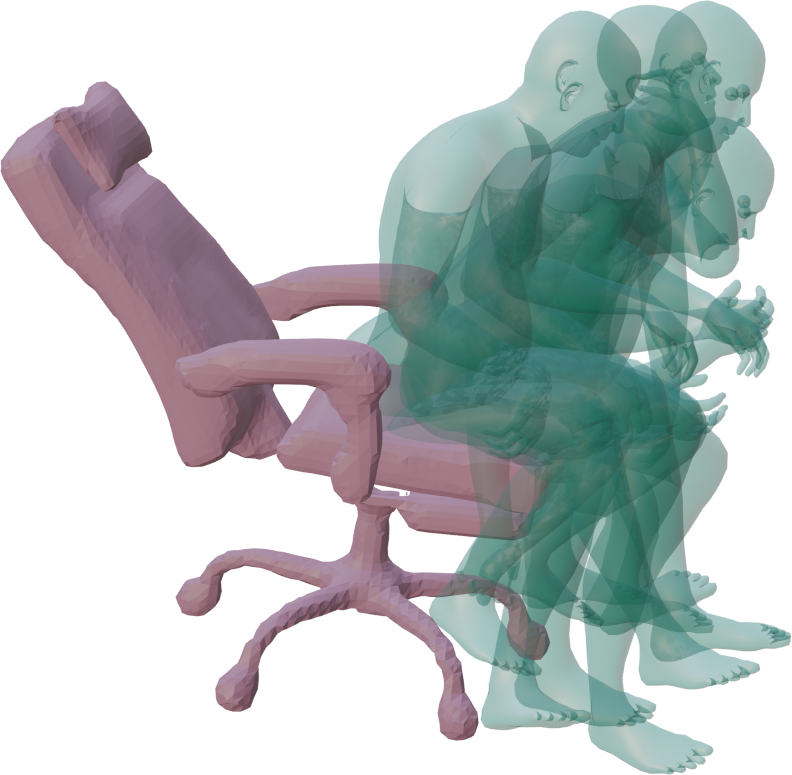}
    \end{subfigure}
    \caption{\textbf{Generated interacting human poses.} Top: model is trained on \ac{dataset}; bottom: model is trained on COUCH~\cite{zhang2022couch}.}
    \label{fig:supp:generative_model}
\end{figure*}

\clearpage

\begin{figure*}[t!]
    \centering
    \includegraphics[width=0.75\linewidth]{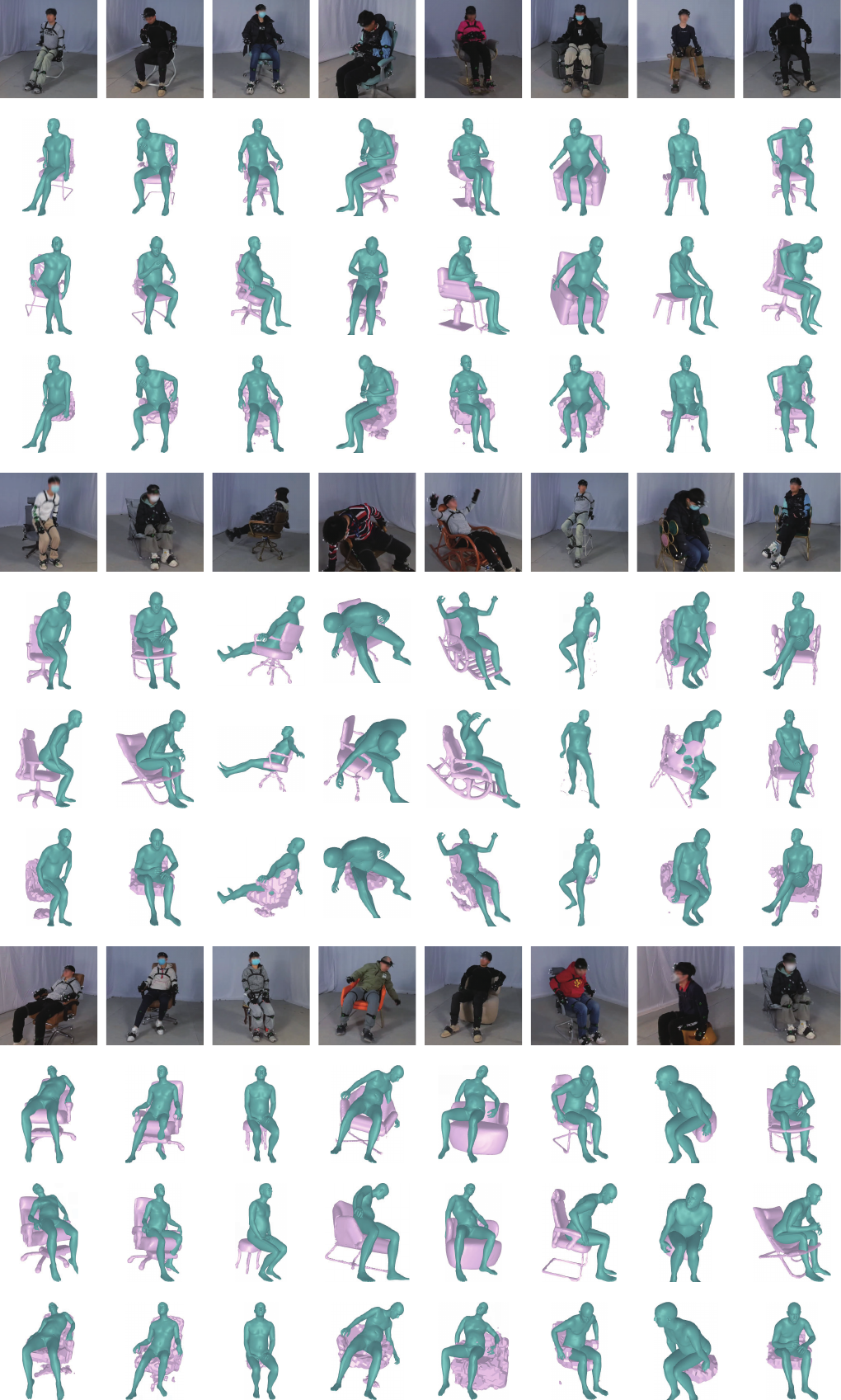}
    \caption{\textbf{Additional qualitative results of our model on the test set of \ac{dataset}.} 
    }
    \label{fig:supp:result}
\end{figure*}

\clearpage

\begin{figure*}[t!]
    \centering
    \includegraphics{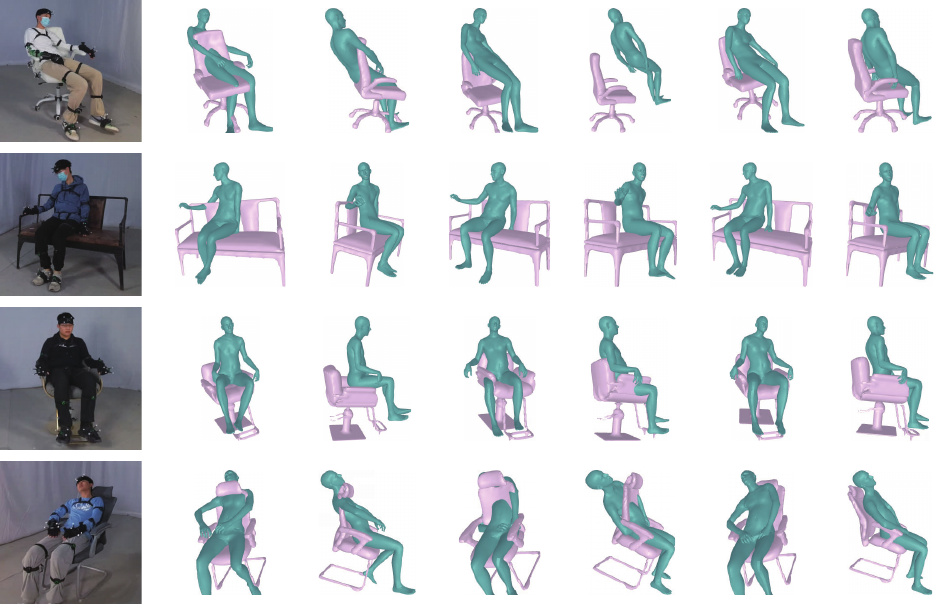}
    \caption{\textbf{Qualitative comparisons.} From left to right: RGB image, CHORE reconstruction, CHORE reconstruction from second view, PHOSA reconstruction, PHOSA reconstruction from second view, \textbf{our reconstruction}, and \textbf{our reconstruction from second view}. Results show a clear advantage of our method in modeling interactions.}
    \label{fig:supp:qual-comp}
\end{figure*}

\clearpage

\begin{figure*}[t!]
    \centering
    \includegraphics[width=\linewidth]{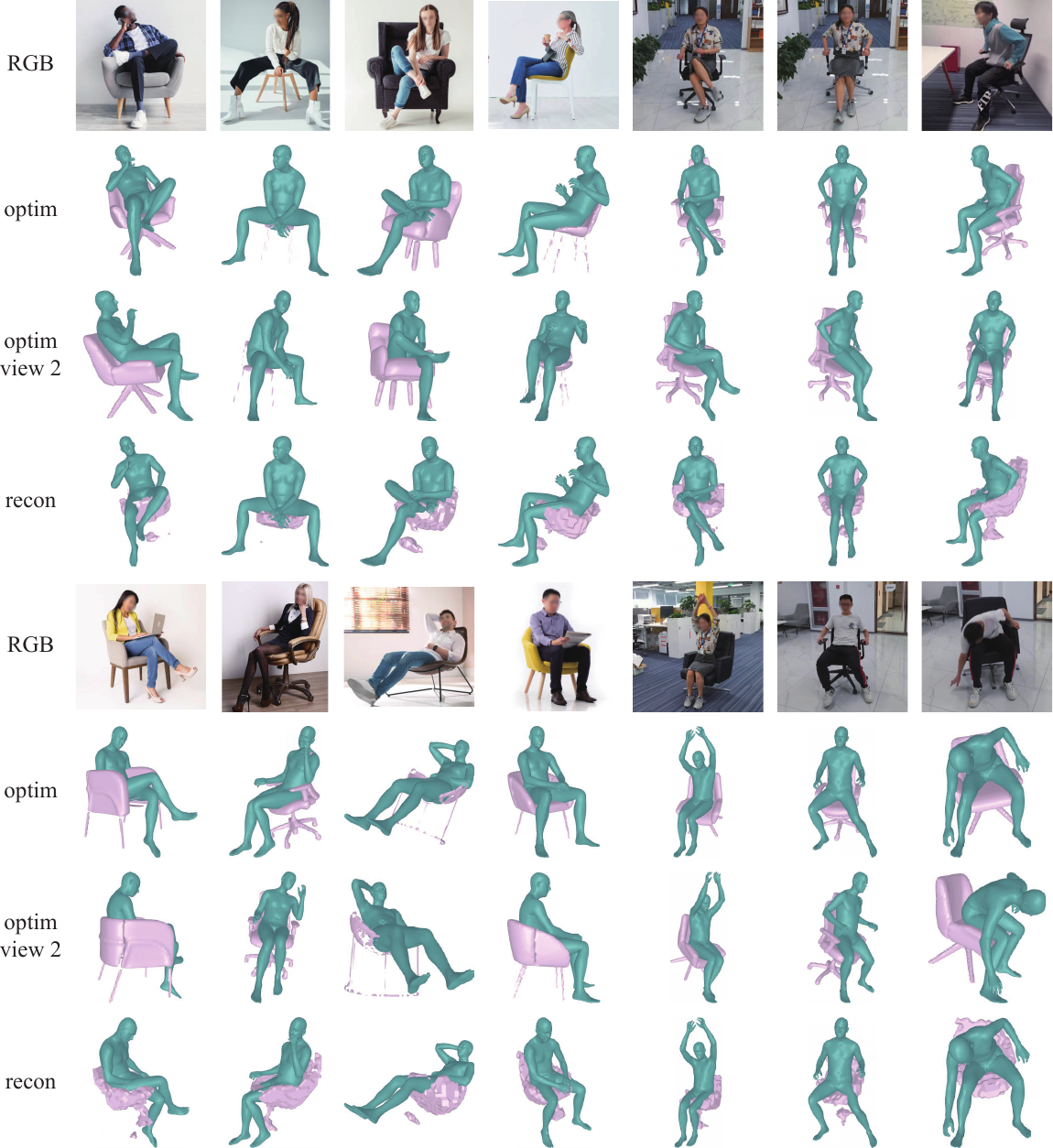}
    \caption{\textbf{Qualitative results of running our model on images captured in the wild.}}
    \label{fig:supp:wild}
\end{figure*}

\clearpage

\begin{figure*}[t!]
    \centering
    \includegraphics[width=\linewidth]{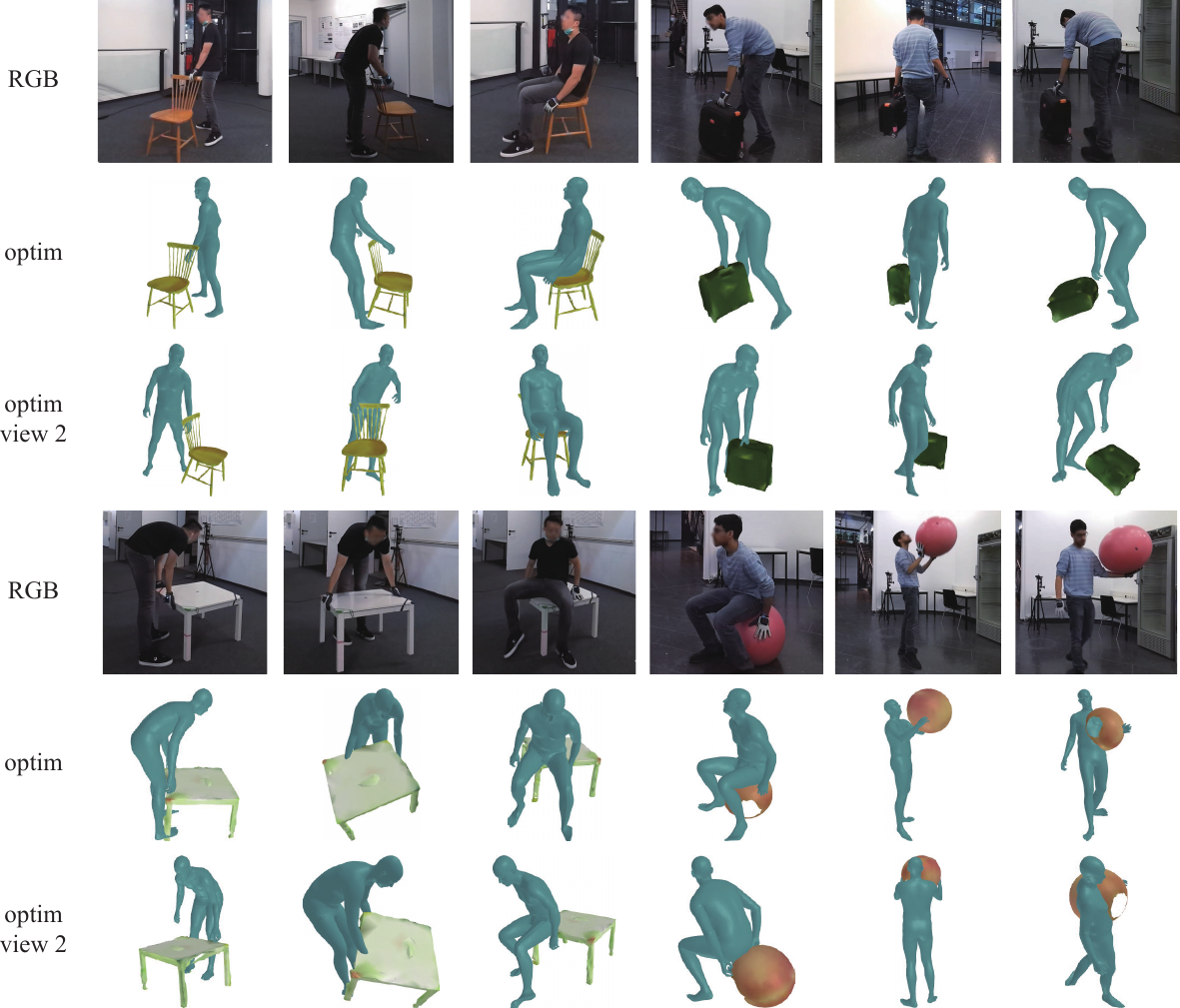}
    \caption{\textbf{Qualitative results of running our model on images from the BEHAVE \cite{bhatnagar2022behave} dataset.}}
    \label{fig:supp:behave}
\end{figure*}

\clearpage

\begin{figure*}[t!]
    \centering
    \includegraphics[width=\linewidth]{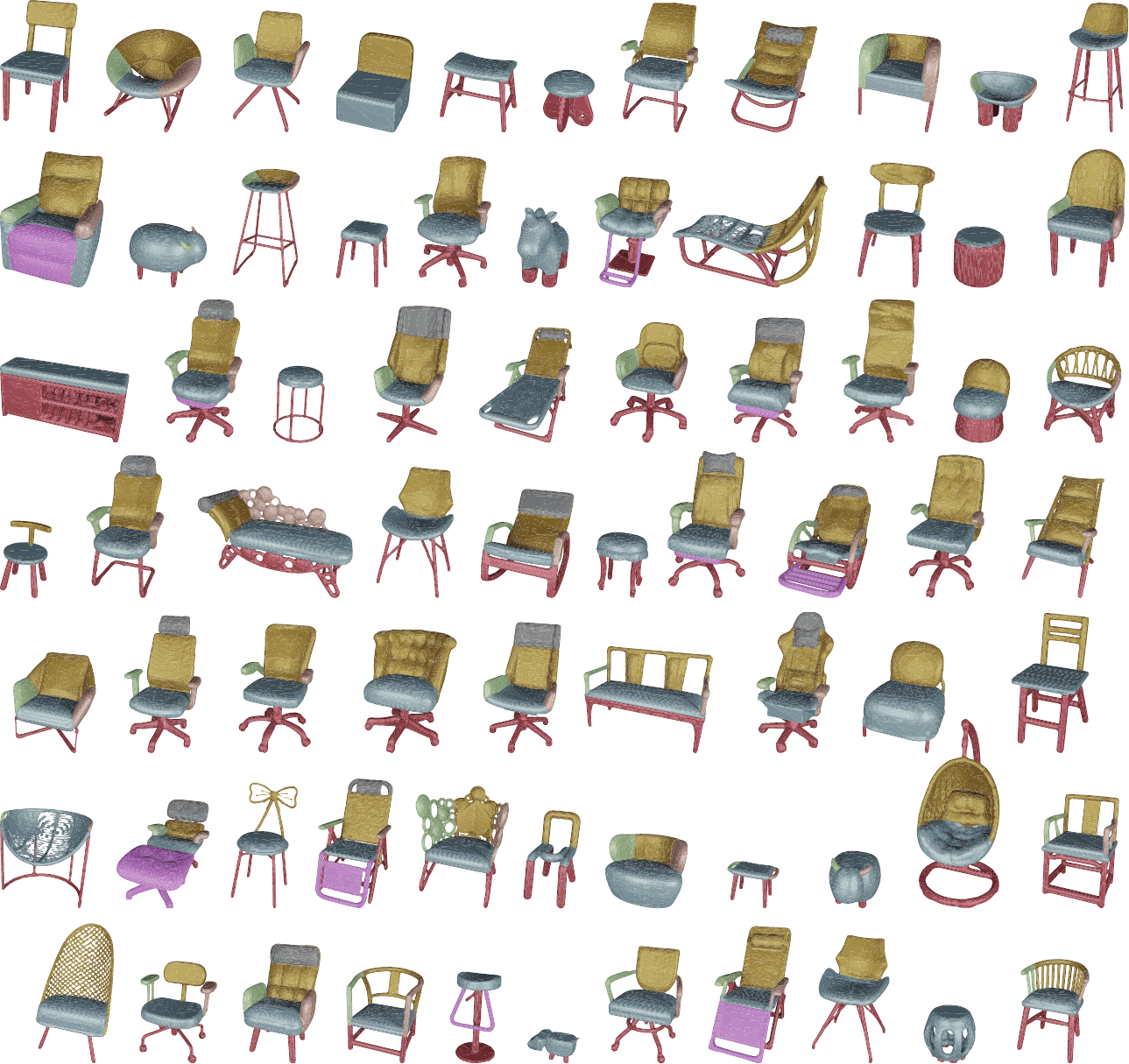}
    \caption{\textbf{Sittable objects in \ac{dataset}.} The first six rows are the objects in the training set, whereas the last row shows the ones in the test set.}
    \label{fig:supp:chairs}
\end{figure*}

\clearpage

\begin{figure*}[t!]
    \centering
    \includegraphics[width=\linewidth]{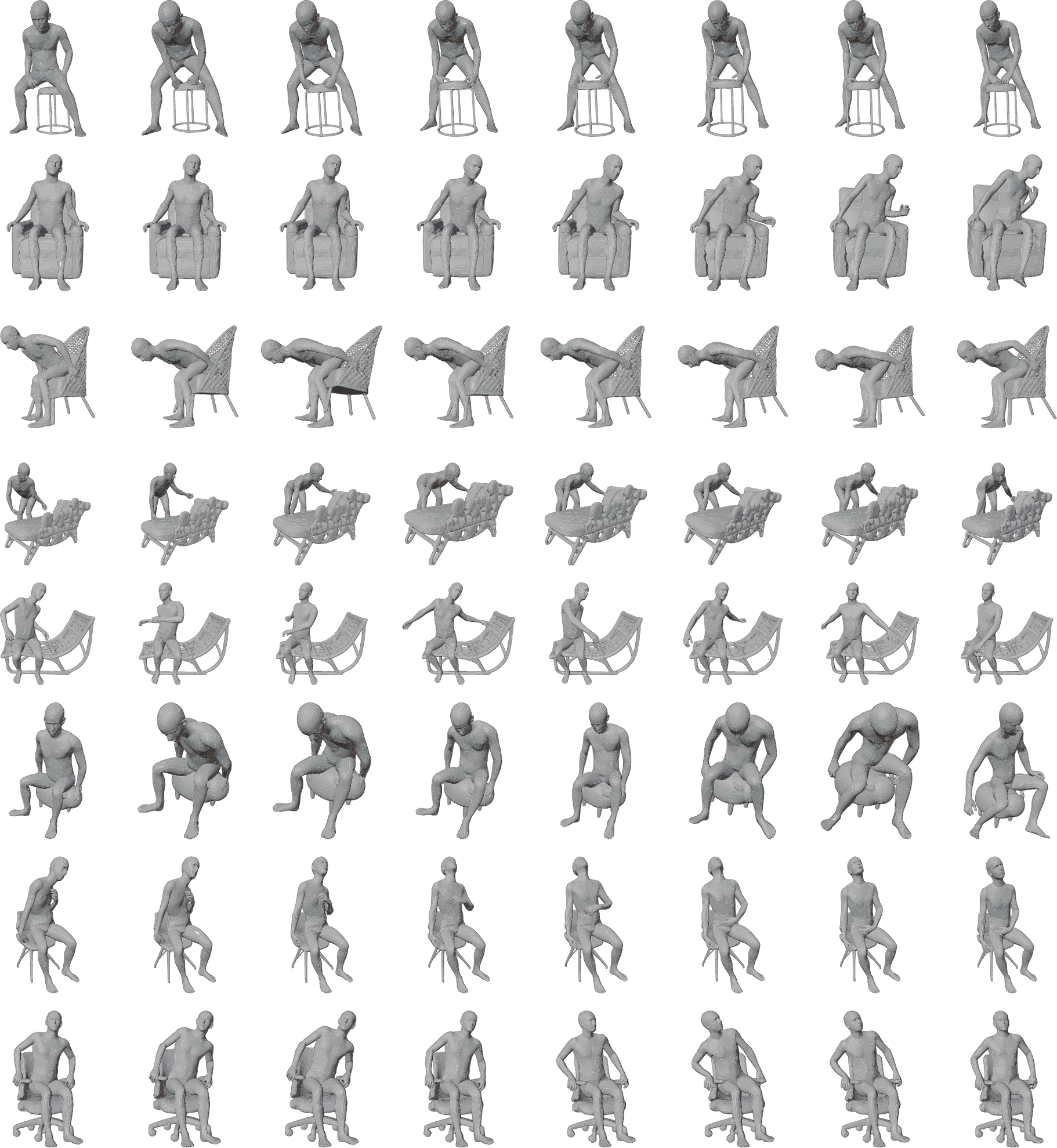}
    \caption{\textbf{Performances of different participants on different objects with the same instruction.} The first four rows show four performances of the instruction ``Move the chair.'' The second participant rotated the chair with a small angle. The last four rows show four performances of the instruction ``Stretch a little in the chair.''}
    \label{fig:supp:trajectories}
\end{figure*}

\end{document}